\documentclass[journal]{IEEEtran}
\usepackage[utf8]{inputenc}

\usepackage[ruled,vlined]{algorithm2e}
\usepackage{amsmath,amssymb,bm}
\usepackage{array,threeparttable,booktabs,multirow}
\usepackage{graphicx}
\usepackage{caption}
\usepackage{subcaption} % load AFTER caption if you’re on IEEEtran
\usepackage{cite}
\usepackage[dvipsnames]{xcolor}
\usepackage{setspace}
\usepackage[colorlinks=true, allcolors=blue]{hyperref}
\usepackage{pifont}
\usepackage{makecell}
\usepackage{diagbox}
\usepackage{setspace}
\usepackage{float}   
\usepackage{enumitem}
\usepackage{orcidlink}
\usepackage{booktabs}         % \toprule, \midrule, \bottomrule
\usepackage{siunitx}          % \textbf, \textbf and unit macros
\usepackage{tcolorbox}
\usepackage[normalem]{ulem}

\usepackage{tikz}

\usepackage{pifont}

\usepackage{booktabs}
\usepackage{multirow}
\usepackage{tabularx}
\usepackage{ragged2e} % For \RaggedRight alignment
\usepackage{pifont}   % For \cmark checkmarks if you use them
\usepackage{amssymb}  % For symbols
\usepackage{tabularx}

\DeclareUnicodeCharacter{FF0C}{,}

\usepackage{tikz}
\usetikzlibrary{shapes, arrows.meta, positioning, calc, fadings, decorations.pathreplacing, patterns, shadows.blur, 3d}
\definecolor{strategicPurple}{RGB}{112, 48, 160} % For HAPS Meta-Control
\definecolor{tacticalGreen}{RGB}{0, 176, 80}    % For BS Data Links
\definecolor{motionOrange}{RGB}{237, 125, 49}   % For UAV Motion vectors
\definecolor{groundGray}{RGB}{242, 242, 242}    % Ground plane
\definecolor{airBlue}{RGB}{220, 230, 245}       % Airspace
\title{
Hierarchical LLM-Driven Control for HAPS-Assisted UAV Networks: Joint Optimization of Flight and Connectivity
}

\author{
Zijiang~Yan\orcidlink{0000-0002-7959-8329}, ~\IEEEmembership{Graduate Student Member,~IEEE},  
Hao~Zhou\orcidlink{0000-0002-5511-4609},~\IEEEmembership{Member,~IEEE},\\  
Wael~Jaafar\orcidlink{0000-0003-4378-9999},~\IEEEmembership{Senior Member,~IEEE},
Jianhua~Pei\orcidlink{0000-0002-4066-9230}, ~\IEEEmembership{Member,~IEEE},
Ping~Wang\orcidlink{0000-0002-1599-5480},~\IEEEmembership{Fellow,~IEEE},
Halim~Yanikomeroglu\orcidlink{0000-0003-4776-9354},~\IEEEmembership{Fellow,~IEEE},
and Hina~Tabassum\orcidlink{0000-0002-7379-6949}~\IEEEmembership{Senior Member,~IEEE}\\ %~\IEEEmembership{Senior Member,~IEEE}

\thanks{
Z. Yan, J. Pei, P. Wang, and H. Tabassum  are with the Department of Electrical Engineering and Computer Science, York University, Toronto, ON, Canada 
(e-mails: \{\href{mailto:zijiang@yorku.ca}{zijiang}, \href{mailto:pingw@yorku.ca}{pingw}, \href{mailto:hinat@yorku.ca}{hinat}\}@yorku.ca, \href{mailto:jianhuapei98@gmail.com}{jianhuapei98@gmail.com} ).
H. Zhou is with Samsung Research America, Toronto, ON, Canada.  (e-mail: \href{mailto:haozhou029@gmail.com}{haozhou029@gmail.com})
W. Jaafar is with the Department of Software and IT Engineering, École de technologie supérieure (ÉTS), University of Quebec, Montreal, QC, Canada  (e-mail: \href{mailto:wael.jaafar@etsmtl.ca}{wael.jaafar@etsmtl.ca}).
H. Yanikomeroglu is with the Non-Terrestrial Networks (Carleton-NTN) Lab and the Department of Systems and Computer Engineering, Carleton University, Ottawa, ON, Canada (e-mail: \href{mailto:halim@sce.carleton.ca}{halim@sce.carleton.ca}).

}
}
\begin{document}
% -------------------------------------------------------------------------
% ABSTRACT & KEYWORDS
% -------------------------------------------------------------------------
\maketitle

\begin{abstract}
% Uncrewed aerial vehicles (UAVs) have been widely adopted in various real-world applications, however, the control and optimization of multi-UAV systems remain a significant challenge, particularly in dynamic and constrained environments. In this context, we investigate the joint flight control and connectivity of multiple UAVs operating within an integrated terrestrial and non-terrestrial network (ITNTN) infrastructure that involves a terrestrial cellular network and a high-altitude platform stations (HAPS). {Specifically}, we consider an aerial highway scenario in which UAVs accelerate, decelerate, and change {positions} to avoid collisions and maintain overall traffic flow. Different from existing studies, we propose a novel hierarchical and collaborative command-and-control (C\&C) method based on large language models (LLMs). Specifically, an LLM deployed on the HAPS performs UAV access control, while other LLMs deployed on UAVs handles detailed motion planning and control. This LLM-based framework leverages the rich knowledge embedded in pre-trained models to enable high-level strategic planning and low-level tactical decisions, revealing its potential for the development of next-generation {three-dimensional (3D)} aerial highway systems. Experimental results demonstrate that our proposed collaborative LLM-based approach achieves higher system rewards, lower operational costs, and significantly reduced UAV collision rates, compared to baseline methods.

Uncrewed aerial vehicles (UAVs) are increasingly deployed in complex networked environments, yet the joint optimization of multi-UAV motion control and connectivity remains a fundamental challenge. In this paper, we study a multi-UAV system operating in an integrated terrestrial and non-terrestrial network (ITNTN) comprising terrestrial base stations and high-altitude platform stations (HAPS). We consider a three-dimensional (3D) aerial highway scenario where UAVs must adapt their motion to ensure collision avoidance, efficient traffic flow, and reliable communication under dynamic and partially observable conditions. We first model the problem as a hierarchical multi-objective partially observable Markov decision process (H-MO-POMDP), capturing the coupling between control and communication objectives. Based on this formulation, we propose a large language model (LLM)-driven hierarchical multi-rate control framework. At the global level, an LLM-based controller on the HAPS performs long-term planning for load balancing and handover decisions. At the local level, each UAV employs a hybrid controller that integrates a slow-timescale LLM for high-level spatial reasoning with a reinforcement learning agent for faster UAV-to-infrastructure (U2I) communication and motion control. We further develop a high-fidelity 3D simulation platform by integrating the gym-pybullet-drones environment with 3GPP-compliant RF/THz channel models. 
% Numerical results show that the proposed framework significantly outperforms state-of-the-art  baselines in terms of system reward, collision reduction, and handover stability, while demonstrating strong zero-shot generalization in  dynamic scenarios.
Numerical results demonstrate that the proposed framework significantly outperforms state-of-the-art baselines, achieving a 14\% increase in transportation efficiency and a 25\% improvement in telecommunication throughput. Additionally, it achieves a 23\% reduction in physical collision rates, demonstrating strong handover stability and zero-shot generalization in dynamic scenarios.

\end{abstract}


\section{Introduction}

% \subsection{Background}
\IEEEPARstart{U}{ncrewed} aerial vehicles (UAVs) are considered as a core component of next-generation Internet-of-Things (IoT) and intelligent transportation systems due to their mobility, flexible deployment, decreasing production costs, and line-of-sight (LoS) connectivity~\cite{yu2022deep}. 
In cellular-connected configurations, UAVs can either act as user equipment (UAV-UEs) that require reliable command-and-control (C\&C) links, or as aerial base stations (BSs) that provide coverage and data services to ground and aerial users. 
Safe and efficient operation of beyond-visual-line-of-sight UAVs critically depends on both wireless connectivity and advanced motion control, especially when UAVs operate cooperatively in shared aerial corridors \cite{cherif20213d,yan2025hierarchical}.

Meanwhile, the integration of High Altitude Platform Stations (HAPS) into 6G networks becomes a promising enabler of ubiquitous, resilient, and cost-effective connectivity and computing services for  users such as UAVs~\cite{JaafarHAPSITS,ren2022caching,Rzig2025}. 
By combining the wide-area coverage and high payload capabilities of HAPS with dense terrestrial BS deployments, integrated terrestrial and non-terrestrial networks (ITNTNs) provide a natural platform for supporting large-scale low-altitude UAV operations. 
In particular, low-altitude UAVs are envisioned to support urban air mobility, cargo delivery, and aerial monitoring, all of which demand not only safe and efficient trajectory control but also high-quality-of-service (QoS) links for C\&C and mission data~\cite{yan2023multi}. 
In such scenarios, UAVs must navigate a three-dimensional (3D) aerial highway where they accelerate, decelerate, and change {positions} to avoid collisions and maintain traffic flow, while simultaneously managing handovers (HOs) and comm in the multi-tier ITNTN~\cite{cherif20213d,Cherif2024}. 
This tight coupling between control and connectivity motivates joint motion--communication design for low-altitude economy.

\subsection{Related Work and Research Gaps}
Research on multi-UAV systems spans motion planning, communication-aware control, HAPS-assisted networking, and, more recently, LLM-driven coordination. Although these directions have progressed independently, they remain insufficiently integrated for dense aerial highway environments. In what follows, we provide a unified discussion that connects these strands and highlights the unresolved challenges that motivate this work.

Early efforts on 3D aerial highway motion planning primarily explored collision avoidance and energy-efficient routing. Applications range from surveillance to factory logistics and precision agriculture~\cite{cherif20213d,kotarski2020design,chen2021bdfl}. A series of studies investigated the combination of reinforcement learning (RL) and convex optimization for safe navigation. For instance, Li \textit{et al.} modeled flying time and energy under kinematic constraints but did not incorporate multi-UAV collision interactions~\cite{li20213d}. Other works adopted multi-agent RL, such as Multi-Agent Deep Deterministic Policy Gradient (MADDPG), for pursuit–evasion~\cite{zhang2022game} or distributed velocity-aware $A^*$-based planning~\cite{hu20203d}. Meanwhile, studies on cargo-UAV energy consumption~\cite{cherif20213d,cherif2021disconnectivity} and speed-controlled data collection~\cite{li2020novel} omitted mutual interference and large-scale traffic dynamics~\cite{yasin2020unmanned,yan2023multi}. As a result, although these works contribute essential foundations, they generally overlook realistic acceleration, deceleration, and dense-lane interactions, leaving robust collision avoidance under high mobility unresolved.

In parallel, communication-centric research examined how UAV movement affects link quality, often assuming simplified or pre-defined trajectories. RL approaches have been used to optimize cell association  and HO metrics~\cite{cherif2023rl,yan2025hierarchical}, and deep learning has been applied to predict HOs in millimeter wave (mmWave) UAV links~\cite{chen2020efficient,yan2025hierarchical}. However, these methods frequently relied on tabular or low-complexity RL variants with slow convergence~\cite{cherif2021disconnectivity,yan2025cvar} and did not incorporate real-time motion dynamics such as lane changes or acceleration. Although recent studies explored speed-HO-traffic tradeoffs~\cite{10077729}, most schemes still treated mobility and communication separately, ignoring HO disruptions triggered by rapid UAV maneuvers.

Beyond terrestrial networks, HAPS platforms have emerged as promising enablers of wide-area UAV coverage. Existing literature demonstrates their potential for HO-aware computing, resource allocation, and latency reduction~\cite{ren2023handoff,chen2025dynamic,Rzig2025}. Yet, these solutions remain largely insensitive to the high-speed motion patterns of UAV traffic and the resulting intense HO dynamics. Furthermore, scalarization-based multi-objective designs failed to capture the coupled nature of transportation and communication constraints. Recent interest in applying LLMs to UAV networking~\cite{yan2025hierarchical,yan2025hybrid,JaafarHAPSITS} pointed toward more adaptive decision-making. Nevertheless, a comprehensive hierarchical framework unifying motion safety, communication reliability, and HAPS coordination is still missing.

Recently, LLM-assisted UAV operations have expanded rapidly due to advances in zero-shot reasoning and semantic understanding. Several works translated natural-language task descriptions into executable flight trajectories using multimodal LLMs integrated with {Robot Operating System (ROS)}~\cite{sutra2025quadcopter}. Others used LLMs for mission adaptation~\cite{tagliabue2024real} or dynamic obstacle avoidance~\cite{baig2026llm}. Nevertheless, these approaches treated motion purely as spatial navigation, without considering how high-speed mobility affects traffic flow or communication stability in multi-UAV settings~\cite{chen2025robust}. 
LLM-based resource management exhibits similar limitations~\cite{zhang2025design}. Indeed, although in-context scheduling~\cite{emami2025llm} and hierarchical cloud–edge LLM coordination~\cite{yan2023multi,han2025research} offer improved autonomy, they typically decouple communication from motion and ignore HO-induced link volatility. 

Finally, larger-scale surveys~\cite{emami2026large,yuan2026enhancing} outlined conceptual possibilities for LLM-enhanced aerial networks, but they lacked concrete formulations or deployable architectures that jointly balance safety, throughput, and communication reliability \cite{yan2025hybrid}.

To summarize, existing research provides valuable progress across motion planning, communication, HAPS coordination, and LLM-enabled control. However, these domains have evolved in isolation, leaving critical gaps in unified joint optimization for dense aerial highways. Motivated by these limitations, we next identify the key challenges that a holistic solution must address.

\subsection{Key Challenges}

Despite substantial progress in UAV networking, motion planning, and HAPS-assisted architectures, designing a unified framework for HAPS-enabled low-altitude UAV systems remains fundamentally challenging. We summarize the main challenges addressed in this work as follows:

\subsubsection{Robust Collision Avoidance Under High Mobility and Dense Traffic}
Low-altitude UAVs increasingly operate in shared and congested aerial corridors, where maintaining safe separation under high mobility is critical. Unlike ground vehicles, UAVs move in 3D space and experience rapid changes in relative locations due to acceleration, deceleration, lane changes, and altitude adjustments~\cite{yan2022reinforcement,yan2023multi}. As UAV density grows, collision risk increases when decisions are made myopically or without coordination~\cite{yasin2020unmanned}. Existing trajectory planning and RL methods often rely on simplified motion models that fail to preserve safety and throughput in dynamic aerial highways~\cite{li20213d}. Standard deep RL (DRL) agents also lack explicit mechanisms for enforcing high-level semantic and regulatory constraints, yielding opaque and difficult-to-certify policies~\cite{yan2023multi,yan2022reinforcement}.

\subsubsection{Communication Reliability and Handover Disruptions}
Reliable wireless connectivity is necessary for beyond-visual-line-of-sight  UAV operations, enabling both C\&C and mission-critical data transfer \cite{yang2025energy}. However, high mobility induces frequent transitions across terrestrial BS footprints and HAPS beams, generating recurrent HO events. Also, poorly timed HOs cause transient link failures, increased latency, and reduced throughput~\cite{cherif2023rl,yan2022reinforcement}. Existing approaches often optimize communication or mobility in isolation, ignoring their mutual dependence~\cite{chen2020efficient}.

\subsubsection{Balancing Transportation Efficiency and Communication Performance}
A fundamental tradeoff exists between transportation efficiency and communication reliability in UAV networks. Indeed, aggressive mobility enhances throughput but increases HO frequency and link instability. Conversely, conservative mobility improves connectivity but degrades transportation efficiency~\cite{10077729}. Fixed-weight scalarization methods cannot adapt to varying network or traffic conditions, often yielding suboptimal performance.

\subsection{Contributions}
Our prior work~\cite{yan2023multi} demonstrated joint motion--communication optimization using branching deep Q-networks~\cite{tavakoli2018action}, but revealed limitations of single-objective DRL in safety-critical, multi-objective ITNTN scenarios. 
{In this work, we address} the identified gaps through a LLM-driven hierarchical control framework that combines HAPS-level LLM meta-control with UAV-level LLM agents. The main contributions of this paper are summarized as:
\begin{itemize}
    \item \textit{A Hierarchical Multi-Objective POMDP (a.k.a, H-MO-POMDP) for Joint Motion--Communication Optimization:} 
    We formalize the tightly coupled problem of 3D autonomous flight and handover-aware connectivity as a hierarchical multi-objective Partially Observable Markov Decision Process (POMDP). 
    The model explicitly embeds the competing goals of collision-free navigation, high throughput traffic flow, and robust telecommunication reliability under realistic partial observability, thereby overcoming the limitations of existing formulations that treat motion and communication as decoupled sub-problems.

    \item \textit{A Generative-AI-Driven Multi-Rate Control Architecture:} 
    We introduce a novel two-tier collaborative LLM-based control framework that transcends fixed-weight RL approaches. 
    At the global tier, a HAPS-deployed meta-controller executes long-horizon strategic planning for load balancing and vertical handover decisions through a reflective reasoning loop. 
    At the tactical tier, each UAV uses a hybrid multi-rate controller in which a slow-timescale LLM provides semantic 3D spatial reasoning and intent shaping, while a fast-timescale double deep Q-network (DDQN), coupled with a motion decoder, performs real-time {UAV-to-infrastrucutre (U2I)} channel selection and continuous C\&C.

    \item \textit{A High-Fidelity 3D ITNTN Simulation and Demonstration of Generalization:} 
    We develop a rigorous 3D testbed by integrating {high-fidelity \textit{gym-pybullet-drones} environment\footnote{An open-source multi-rotor physics simulator based on PyBullet, utilized for our experimental evaluations in Section~\ref{sec:numerical_results}. Source code: \url{https://github.com/utiasDSL/gym-pybullet-drones}} \cite{panerati2021learning}}
    with standardized 3GPP RF/THz propagation models, departing from the oversimplified 2D/abstract analytic environments commonly used in prior work. 
    Extensive numerical experiments show that the proposed hierarchical architecture consistently outperforms state-of-the-art DRL and Multi-Agent RL (MARL) baselines in overall reward, collision mitigation, and handover stability, while exhibiting robust zero-shot adaptability in dense and rapidly evolving aerial highway scenarios.
\end{itemize}

The remainder of this paper is organized as follows. Section~\ref{systemmodel} presents the ITNTN system model, detailing both the multi-tier communication infrastructure and the high-fidelity UAV physical kinematics. Section~\ref{sec:mdp_formulation} formulates the joint flight navigation and network association problem as a hierarchical multi-objective POMDP (H-MO-POMDP). Section~\ref{sec:algorithm} introduces the proposed generative AI-driven collaborative framework, outlining the cognitive reasoning loops for the HAPS meta-controller and the UAV tier edge-agents. Section~\ref{sec:numerical_results} provides comprehensive numerical results  to validate the effectiveness of the proposed architecture. Finally, Section~\ref{sec:conclusion} concludes the paper.

% -------------------------------------------------------------------------
% SYSTEM MODEL
% -------------------------------------------------------------------------

\begin{figure}[t]
    \centering
    \includegraphics[trim={1.2cm 0 0.7cm 0},clip,width=1\linewidth]{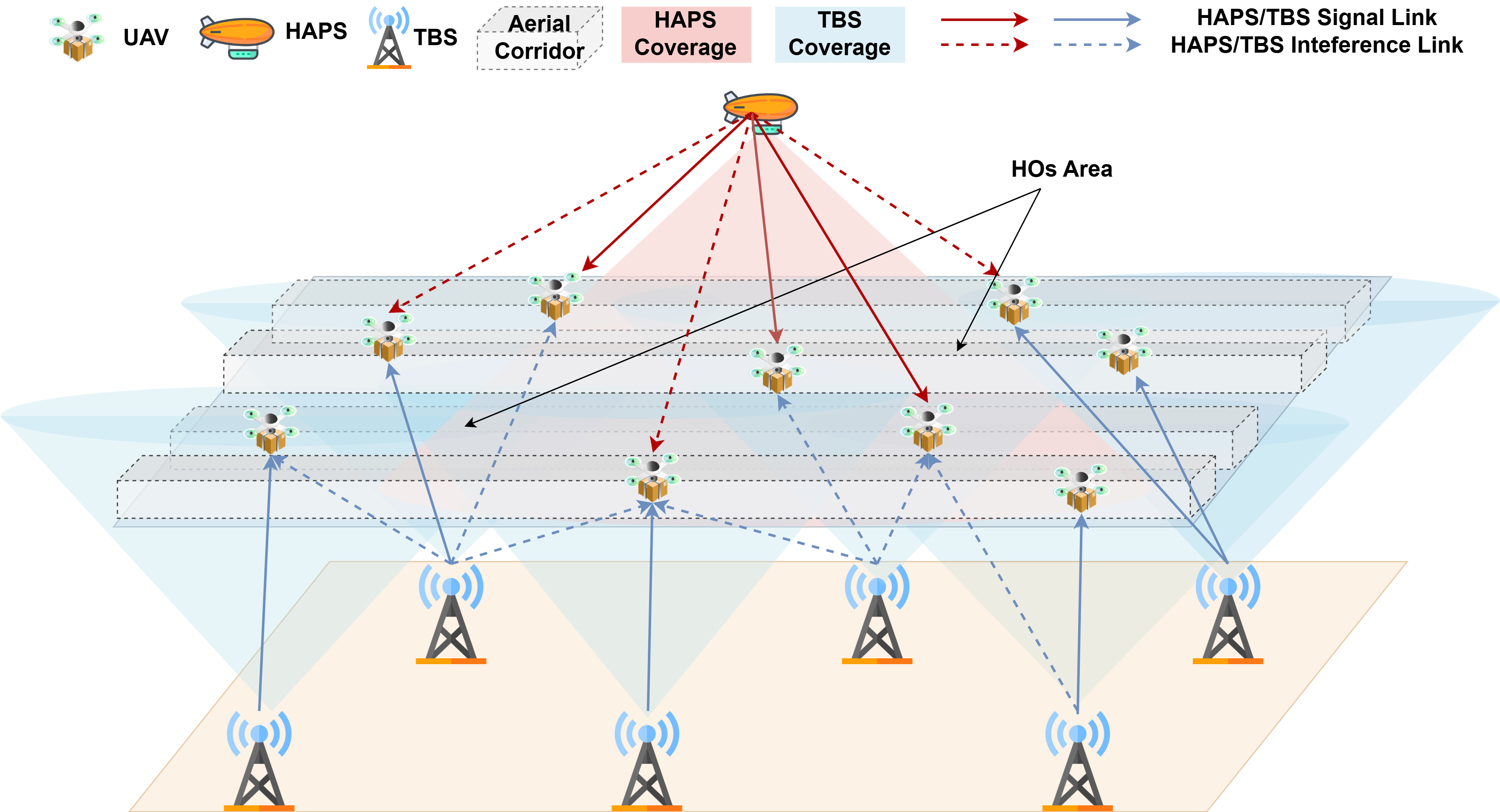}
    \caption{3D HAPS-assisted UAV network model. %The downward-pointing red cones represent the wide-area coverage spot-beams provided by the HAPS. The upward-pointing blue cones indicate the localized coverage areas of the BSs. Solid/dashed lines represent desired/interference links. 
    %% The UAVs navigate shared aerial corridors, necessitating the joint optimization of physical flight kinematics and handover-aware multi-tier communication.
    }  
\label{fig:bdq_agent}
\end{figure}

\section{System Model}
\label{systemmodel}
{
As illustrated in Fig. 1, we consider a dynamic 3D aerial highway scenario where a set of $M$ UAVs, denoted by $\mathcal{M}= \{1, 2, \dots ,M\}$, operate within an ITNTN infrastructure. This infrastructure comprises a set of $B$ terrestrial base stations (TBSs), denoted by $\mathcal{B} = \{1, 2, \dots ,B\}$, and a high-altitude platform station (HAPS) node denoted by $H$. 

% To explicitly clarify the operational scope of this network,
The specific mission and communication requirements of the UAVs are defined as follows:

\textbf{(a) Heterogeneous 3D Transit Missions:} Rather than acting as a synchronized swarm performing a single collaborative task, each UAV $m \in \mathcal{M}$ is tasked with an independent logistics or transit mission. Each UAV must navigate from its unique origin to a specific 3D destination ($\mathbf{x}_{\text{target}}^m$). The UAVs are not constrained to a uniform altitude, they must dynamically exploit the full continuous 3D airspace to avoid collisions within the shared corridor.

\textbf{(b) Mission-Critical Communication Constraints:} Operating safely in a high-density, high-speed aerial corridor requires uninterrupted command-and-control (C2) and telemetry links. A communication outage or severe latency spike directly compromises physical flight safety. Therefore, the UAVs must maintain continuous, high-quality data rates ($\text{WR}_t^{m,c}$) to stream vital sensor data and receive traffic coordination directives from the HAPS meta-controller.

Because UAVs traverse the overlapping coverage areas of the HAPS and varying TBSs at high velocities, they inevitably trigger network handovers. Consequently, this system model encompasses two highly coupled and dynamic domains, namely the multi-tier communication architecture that dictates link reliability and load balancing, and the continuous physical kinematics that govern real-time UAV flight safety and aerodynamic stability.
}

\subsection{Multi-Tier Communication Model}

\subsubsection{Ground-to-Air (G2A) Channel Model}
The G2A channel model between a terrestrial BS $b \in \mathcal{B}$ and a UAV $m \in \mathcal{M}$ is fundamentally dependent on the 3D distance $d_t^{m,b}$ between them at time slot $t$, the BS antenna gain, and the LoS probability \cite{3gpp777}. In cellular-connected aerial systems, UAVs frequently rely on the sidelobe emissions of BS antennas. We adopt the 3GPP antenna pattern specification \cite{3gpp777}, partitioning each BS into three sectors with cross-polarized elements in a uniform linear array (ULA). 

Each antenna element delivers a peak gain of $B_{\max}=8$ dBi along its principal lobe. The sidelobe gains vary according to the azimuth angle $\phi^{m,b}_t$ and elevation angle $\zeta^{m,b}_t$ between BS $b$ and UAV $m$ at time $t$ \cite{cherif2022cellular}:
\begin{equation}
    B_{\mathrm{az}}(\phi^{m,b}_t) =\min \left\{ 12 \left( \frac{\phi^{m,b}_t}{\phi_\mathrm{3dB}} \right)^2, \mathrm{B_m}  \right\},
\end{equation}
% \begin{equation}\label{eq:g_az}
%     G_{\mathrm{az}}(\phi_t^{m,b}) = \min\left\{ 12 \left( \frac{\phi_t^{m,b}}{\phi_{\mathrm{3dB}}=65^\circ} \right)^2, G_{\mathrm{m}} = -17.2 \right\},
% \end{equation}
\begin{equation}
    B_{\mathrm{el}}(\zeta^{m,b}_t) =\min \left\{ 12 \left( \frac{\zeta^{m,b}_t}{\zeta_\mathrm{3dB}} \right)^2, \mathrm{SLA}  \right\},
\end{equation}
where $\phi_\mathrm{3dB}=\zeta_\mathrm{3dB}=\frac{65\pi}{180}$ represents the 3 dB bandwidths, and $\mathrm{B_m}$ and $\mathrm{SLA}$ are the antenna null thresholds. The combined element gain is:
{\small
\begin{equation}
    B(\zeta^{m,b}_t, \phi^{m,b}_t) = B_{\mathrm{max}} - \min \{- (B_{\mathrm{az}}(\phi^{m,b}_t) + B_{\mathrm{el}}(\zeta^{m,b}_t) ) , \mathrm{B_m} \}.
\end{equation}}
Assuming BS $b$ is equipped with $N$ antennas with a down-tilt $\zeta_b^d$, the array factor is:
\begin{equation}
    \mathrm{F}(\zeta^{m,b}_t) = \frac{\sin\left(\frac{N\pi}{2} (\sin \zeta^{m,b}_t - \sin \zeta_b^d)\right)}{\sqrt{N} \sin\left(\frac{\pi}{2} (\sin\zeta^{m,b}_t - \sin\zeta_b^d)\right)}.
\end{equation}
Then, the total array radiation pattern from BS $b$ to UAV $m$ is $B^{m,b}_t = B(\zeta^{m,b}_t,\phi^{m,b}_t) + \mathrm{F}(\zeta^{m,b}_t)$.

The probability that UAV $m$, flying at altitude $h_m$, maintains a LoS connection with BS $b$ at distance $d_t^{m,b}$ is
\begin{equation}
    P_{\mathrm{LoS}}(d_t^{m,b})= 
\begin{cases}
1,\quad & \text{if } d_t^{m,b} \leq d_1 \\
\frac{d_1}{d_t^{m,b}} + e^{-\frac{d_t^{m,b}}{p_1}}\left(1- \frac{d_1}{d_t^{m,b}} \right),\quad &\text{otherwise,}
\end{cases} 
\end{equation}
where $d_1 = \max\{460\log_{10}(h_m)-700, 18\}$ and $p_1 = 4300\log_{10}(h_m)-3800$. For $h_m \in [100,300]$ m, we assume $P_{\mathrm{LoS}} \approx 1$. 

{
The probabilistic mean path loss $L^{m,b}_t$ (in dB) is calculated as
\begin{equation}
    L^{m,b}_t = L_b^{\mathrm{LoS}}P_{\mathrm{LoS}}(d_t^{m,b}) + L_b^{\mathrm{NLoS}}(1 - P_{\mathrm{LoS}}(d_t^{m,b})),
\end{equation}
where $L_b^{\mathrm{LoS}}$ and $L_b^{\mathrm{NLoS}}$ are the LoS and non-LoS (NLoS) path loss components.

To formulate the signal-to-interference-plus-noise ratio (SINR), we first define the linear channel power gain $G_t^{m,b}$ between UAV $m$ and terrestrial BS $b$ as
\begin{equation}
    G_t^{m,b} = 10^{(G_{\max} - L^{m,b}_t) / 10},
\end{equation}
where $G_{\max}$ is the peak antenna gain in dBi. 

Assuming downlink transmission, the received SINR in linear scale at UAV $m$ is given by
\begin{equation}
\label{eq:sinr_g2a}
    \text{SINR}_t^{m,b} = \frac{P_T G_t^{m,b}}{N_0 B_b + \sum_{b' \in \mathcal{B} \setminus \{b\}} P_T G_t^{m,b'}},
\end{equation}
where $P_T$ denotes the transmission power of the terrestrial base stations in milliwatts (mW), $N_0$ represents the noise power spectral density in mW/Hz, and $B_b$ represents the bandwidth allocated to terrestrial base station $b$ in Hz. The summation term in the denominator represents the aggregate downlink interference received from all other transmitting terrestrial BSs.
}

\subsubsection{UAV-HAPS Channel Model}
We assume that each UAV is equipped with a single antenna, whereas the HAPS utilizes a multi-antenna array to generate spot-beams. Given a total available HAPS bandwidth {$B_{\mathrm{HAPS}}$}, the fraction allocated to UAV \(m\) is \( b_t^{H,m} \in [0,1]\), subject to $   \sum_{m \in \mathcal{M}} b_t^{H,m} \le 1$. The UAV-HAPS link is dominated by LoS conditions, yielding an instantaneous channel gain \cite{alsharoa2020improvement}.
\begin{equation}\label{eq:channel_gain}
    B_{t}^{m,H} = B_H  \biggl(\frac{v_l}{4\pi\,d_{t}^{m,H}\,f_{c}}\biggr)^{2} \bigl|h_t^{m,H}\bigr|^{2},
\end{equation}
where {\(v_l\)} is the speed of light, \(f_{c}\) is the carrier frequency, \(B_H\) is the directional antenna gain, and \(h^{m,H}\) is the small-scale Rician-fading coefficient, and 
$d_t^{m,H}$ is the 3D distance between UAV $m$ and HAPS $H$.
The small-scale fading $h_t^{m,H}$ follows a Rician distribution, i.e., 
\begin{equation}\label{eq:rician}
    h_{t}^{m,H} \sim \frac{1}{\sqrt{2}} \left( \sqrt{\frac{K}{K+1}} e^{j\theta} + \sqrt{\frac{1}{K+1}} (g_I + j g_Q) \right),
\end{equation}
where $K=15$ dB is the Rician factor, $\theta \sim \mathcal{U}[0,2\pi)$ is the LoS phase, $g_I, g_Q \sim \mathcal{N}(0,1)$ {are independent and identically distributed (i.i.d.) Gaussian random variables representing} the in-phase/quadrature scattering components, and $j = \sqrt{-1}$ imaginary unit ~\cite{alsharoa2020improvement}.

Subsequently, the 
%This $B_t^{m,H}$ enters the rate eq.~(\ref{eq:rate}) as SNR numerator.
achievable data rate over the UAV-HAPS link at time $t$ is
\begin{equation}\label{eq:rate}
    R_t^{m,H} = b_t^{H,m} B_{\mathrm{HAPS}} \log_2 \Biggl( 1 + \frac{p_t^{H,m} P_{\max} B_t^{m,H} }{b_t^{H,m} B_{\mathrm{HAPS}} N_0} \Biggr),
\end{equation}
where 
%$B_t^{m,H} = |h_t^{m,H}|^2$ is the instantaneous channel power gain  and 
\(p_t^{H,m} \in [0,1]\) is the dynamic power allocation fraction at time $t$,  $P_{\max}$ is the UAV transmit power limit.

\subsubsection{Weighted Data Rate with Handover Constraints}
Each node $c \in \mathcal{C} = \mathcal{B} \cup \{HAPS\}$ has a finite capacity quota $Q_c$, which is the maximum number of UAV users it can simultaneously serve. Let $n_{c,t}$ be the current number of associated UAV users to node $c \in \mathcal{C}$ at time $t$. The weighted rate of UAV $m$ served by node $c$ penalizes overload and HOs as follows:
\begin{equation}\label{eq:weight_data_rate}
    \mathrm{WR}_t^{m,c} = \frac{R_t^{m,c}}{\min(Q_c, n_{c,t})} (1 - \gamma \mathbb{I}_{\mathrm{HO},t}),\; \forall m \in \mathcal{M},\; c \in \mathcal{C},
\end{equation}
where $R_t^{m,c}$ is given by eq.(\ref{eq:rate}), $\min(Q_c, n_{c,t})$ allocates per-user share under saturation (i.e, $n_{c,t} > Q_c$ slows down new UAV associations to node $c$), $\gamma$ penalizes HO switches where $\mathbb{I}_{\mathrm{HO},t}=1$ if UAV $m$ was associated to a node different from $c$ at time $t-1$, otherwise, $\mathbb{I}_{\mathrm{HO},t}=0$.
This discourages overcrowding the BSs and HAPS. %(e.g., HAPS rejects when $n_{H,t} \ge Q_H$) and HOs.

\subsection{Physical Kinematics and Dynamics Model}
\label{subsec:dynamics_model}

We adopt a high-fidelity dynamic model for the quadcopter UAVs based on the rigid body formulation in \cite{panerati2021learning}. 
%This model captures the coupled rotational and translational dynamics driven by the forces and torques generated by the four rotors.

%\subsubsection{Rigid Body Dynamics}
Let the position of the $m$-th UAV in the inertial frame be $\mathbf{x}_m = [x_m, y_m, z_m]^\top$, and its attitude be represented by the rotation matrix $\mathbf{R}_m$, {which defines transformation} from the body frame to the inertial frame. The translational dynamics are governed by Newton's second law:
\begin{equation}
    \mathbf{\ddot{x}}_m = \frac{1}{M_u} \left( \mathbf{R}_m \cdot \mathbf{F}_{T,m} - \mathbf{F}_g - \mathbf{F}_{\text{drag},m} \right), \; \forall m \in \mathcal{M},
    \label{eq:trans_dynamics}
\end{equation}
where $M_u$ is the UAV mass, $\mathbf{F}_g = [0, 0, M_ug]^\top$ is the gravity vector, $g$ is gravitational constant, and $\mathbf{F}_{T,m} = [0, 0, \sum_{k=1}^4 F_{k,m}]^\top$ represents the total thrust generated by the motors in the body frame. The drag force $\mathbf{F}_{\text{drag},m}$ accounts for air resistance and is modeled proportional to the linear velocity $\mathbf{\dot{x}}_m$ \cite{forster2015}:
\begin{equation}
    \mathbf{F}_{\text{drag},m} = \mathbf{D} \cdot \mathbf{\dot{x}}_m,
\end{equation}
where $\mathbf{D}$ is the diagonal drag coefficient matrix.

The rotational dynamics describe the evolution of the angular velocity $\boldsymbol{\omega}_m$ in the body frame as
\begin{equation}
    \mathbf{J} \dot{\boldsymbol{\omega}}_m = \boldsymbol{\tau}_m - \boldsymbol{\omega}_m \times (\mathbf{J} \boldsymbol{\omega}_m),
    \label{eq:rot_dynamics}
\end{equation}
where $\mathbf{J}$ is the diagonal inertia matrix and $\boldsymbol{\tau}_m = [\tau_{x,m}, \tau_{y,m}, \tau_{z,m}]^\top$ is the torque vector generated by the differential thrust of the rotors.

%\subsubsection{Motor Forces and Torques}
The force $F_{k,m}$ and torque contribution produced by the $k$-th motor ($k \in \{1, \ldots, 4\}$) are proportional to the square of its rotational speed $P_{k,m}$ (in round per minute -RPM) \cite{Powers2015}, given by
% \begin{equation}
% \label{eq:motor_forces}
%     F_{k,m} = k_F P_{k,m}^2, \; M_{k,m} = k_T P_{k,m}^2,\; \textcolor{blue}{\forall} k \in \{1, \ldots,4 \}, \; m \in \mathcal{M},
% \end{equation}
\begin{equation}
\label{eq:motor_forces}
\begin{aligned}
    F_{k,m} &= k_F P_{k,m}^2, 
    \quad M_{k,m} = k_T P_{k,m}^2, \\
    &\forall k \in \{1, \ldots,4 \}, \; m \in \mathcal{M},
\end{aligned}
\end{equation}
where $k_F$ and $k_T$ are the thrust and torque coefficients, respectively, and  $M_{k,m}$ is the yaw moment produced by the spinning rotor. For a standard X-configuration quadcopter, the total body torques are expressed by
\begin{equation}
    \boldsymbol{\tau}_m = 
    \begin{bmatrix}
        \frac{L}{\sqrt{2}} (F_{1,m} - F_{2,m} - F_{3,m} + F_{4,m}) \\
        \frac{L}{\sqrt{2}} (F_{1,m} + F_{2,m} - F_{3,m} - F_{4,m}) \\
        \sum_{k=1}^4 (-1)^{k+1} M_{k,m}
    \end{bmatrix},
\end{equation}
where $L$ denotes the arm length of the UAV.

%\subsubsection{Aerodynamic Effects}
To enhance simulation realism, we incorporate the {ground effect}, which increases lift when the UAV operates close to the surface. Following \cite{Shi2019NeuralLander}, the thrust of the $k$-th motor, called $G_{k,m}$, is augmented by a factor, depending on the altitude $h_{k,m}$ of the propeller, as
% \begin{equation}
%     G_{k,m} = k_G \cdot k_F \left( \frac{r_P}{4 h_{k,m}} \right)^2 P_{k,m}^2, \; \forall k \in \{1, \ldots,4 \}, \; m \in \mathcal{M},
% \end{equation}
\begin{equation}
\begin{aligned}
    G_{k,m} &= k_G \cdot k_F \left( \frac{r_P}{4 h_{k,m}} \right)^2 P_{k,m}^2, \\
    &\forall k \in \{1, \ldots,4 \}, \; m \in \mathcal{M},
\end{aligned}
\end{equation}
where $r_P$ is the propeller radius and $k_G, k_F$ are the ground effect coefficient and {thrust coefficient, respectively.} 
% This term is added to the standard thrust $F_{k,m}$ in (\ref{eq:motor_forces}) when operating in low-altitude scenarios, including the take-off and landing phases.
During low-altitude scenarios, the effective thrust of each  motor is updated to $F_{k,m} + G_{k,m}$, which proportionally increases the total thrust vector $\mathbf{F}_{T,m}$ in Eq. (\ref{eq:trans_dynamics}).

Although the environment can be accurately simulated with precise multi-rotor RPMs and torques, directly optimizing these low-level variables is computationally complex for discrete semantic agents like LLMs. To bypass this issue, our proposed framework delegates only high-level semantic waypoints and velocity targeting to the LLM agent. Subsequently, a lower-level motion decision decoder maps these semantic intentions into the precise continuous rotation speed required by the rigid-body kinematic model, thereby closely coupling the physical reality of flight with the strategic reasoning of generative AI.
% \red{A minor question is that can we specify a detailed optimization problem here? Like at each level, the objective and constraints are xxxx. }
{To capture the coupled dynamics of 3D trajectory control and handover-aware cell association, we first define the global joint optimization problem $\mathcal{P}$. The overarching objective is to maximize the aggregate system utility, encompassing both transportation efficiency and telecommunication data rates, while strictly adhering to physical kinematics, safety distances, and ITNTN capacity constraints.

Let $u_t^{m,c} \in \{0, 1\}$ be a binary association indicator, where $u_t^{m,c} = 1$ if UAV $m$ connects to serving node $c \in \mathcal{C}$ at time $t$, and $u_t^{m,c} = 0$ otherwise. The centralized joint optimization problem is formulated as follows:
\begin{equation}
\label{eq:optimization_problem}
\begin{aligned}
\mathcal{P}: \quad & \max_{\mathbf{A}_{\text{mot}}, \mathbf{U}} \quad \lim_{T \to \infty} \frac{1}{T} \sum_{t=1}^{T} \sum_{m \in \mathcal{M}} \left[ R_{\text{tran},t}^m + \sum_{c \in \mathcal{C}} u_{t}^{m,c} \text{WR}_{t}^{m,c} \right] \\
\text{s. t. \quad}&   \mathrm{C1}: \sum_{c \in \mathcal{C}} u_{t}^{m,c} = 1, \quad \forall m \in \mathcal{M}, \forall t, \\
& \mathrm{C2}: \sum_{m \in \mathcal{M}} u_{t}^{m,c} \leq Q_{c}, \quad \forall c \in \mathcal{C}, \forall t, \\
& \mathrm{C3}: \sum_{m \in \mathcal{M}} u_{t}^{m,H} R_t^{m,H} \leq C_{\max}^{\text{HAPS}}, \quad \forall t, \\
& \mathrm{C4}: \|\mathbf{x}_t^m - \mathbf{x}_t^j\|_2 \geq d_{\text{safe}}, \quad \forall m \neq j \in \mathcal{M}, \forall t, \\
& \mathrm{C5}: \mathbf{x}_{t+1}^m = \Phi(\mathbf{x}_t^m, \mathbf{a}_{\text{mot},t}^m), \quad \forall m \in \mathcal{M}, \forall t, \\
& \mathrm{C6}: u_{t}^{m,c} \in \{0,1\}, \quad \forall m, c, t.
\end{aligned}
\end{equation}
In this formulation, $\mathbf{A}_{\text{mot}}$ represents the continuous motion actions (rotor RPMs) and $\mathbf{U}$ represents the discrete network association matrix. The constraints are defined as follows:
\textbf{C1} Ensures each UAV maintains exactly one active C\&C link with a serving node (TBS or HAPS) at any given time.
%
% textbf{C2 \& C3 (Network Capacity Limits):} 
\textbf{C2} restricts the number of associated users per node to its maximum hardware quota $Q_c$. For the HAPS tier, \textbf{C3} explicitly guarantees that the aggregated data rate does not exceed the total backhaul capacity $C_{\max}^{\text{HAPS}}$, preventing systemic network saturation.
\textbf{C4} is a strict physical safety constraint ensuring that the 3D Euclidean distance between any two UAVs never falls below the safety threshold $d_{\text{safe}}$.
\textbf{C5} Ensures that the continuous spatial evolution $\Phi(\cdot)$ strictly obeys the high-fidelity 3D rigid-body dynamic equations defined in Eq. (\ref{eq:trans_dynamics})--(\ref{eq:motor_forces}), driven by the continuous rotor commands $\mathbf{a}_{\text{mot},t}^m$.
% \begin{itemize}
%     \item \textbf{C1 (Association Uniqueness):} Ensures each UAV maintains exactly one active C\&C link with a serving node (TBS or HAPS) at any given time.
%     \item \textbf{C2 \& C3 (Network Capacity Limits):} C2 restricts the number of associated users per node to its maximum hardware quota $Q_c$. For the HAPS tier, C3 explicitly guarantees that the aggregated data rate does not exceed the total backhaul capacity $C_{\max}^{\text{HAPS}}$, preventing systemic network saturation.
%     \item \textbf{C4 (Collision Avoidance):} A strict physical safety constraint ensuring that the 3D Euclidean distance between any two UAVs never falls below the safety threshold $d_{\text{safe}}$.
%     \item \textbf{C5 (Kinematic Feasibility):} Ensures that the continuous spatial evolution $\Phi(\cdot)$ strictly obeys the high-fidelity 3D rigid-body dynamic equations defined in Eq. (\ref{eq:trans_dynamics})--(\ref{eq:motor_forces}), driven by the continuous rotor commands$\mathbf{a}_{\text{mot},t}^m$.
% \end{itemize}

Problem $\mathcal{P}$ is a highly non-convex mixed-integer non-linear programming problem with deeply coupled continuous physics and discrete associations in high-mobility environments. 
% Obtaining a centralized analytical solution is computationally intractable for real-time control in high-mobility environments.
Therefore, to make the problem solvable, we decompose this optimization via a two-tier Hierarchical Multi-Objective POMDP, delegating specific constraints to distinct LLM intelligent tiers:
\begin{itemize}
    \item \textbf{Global Strategic Level (HAPS Meta-Controller):} The massive cloud-deployed LLM explicitly addresses constraints \textbf{C2} and \textbf{C3}. By operating at a slow timescale, it orchestrates the global association matrix $\mathbf{U}$ to ensure load-balancing and proactively prevent capacity violations across the ITNTN.
    \item \textbf{Local Tactical Level (UAV Edge-Agents):} The lightweight UAV-deployed LLMs alongside the DDQN explicitly handle constraints \textbf{C4} and \textbf{C5}. They continuously optimize the physical thrust actions $\mathbf{a}_{\text{mot},t}^m$ and execute localized U2I handovers to maximize trajectory efficiency and safety, dynamically adapting their objectives based on the strategic network boundaries imposed by the HAPS.
\end{itemize}
}

\section{Hierarchical MO-POMDP Formulation for UAV Control}
\label{sec:mdp_formulation}

To capture the complex interplay between continuous flight dynamics and discrete network associations, we formulate the multi-UAV control problem as a Hierarchical Multi-Objective Partially Observable Markov Decision Process (H-MO-POMDP). In practical simulation environments, agents rarely have access to the perfect global state; instead, they must rely on localized, partial observations. Within this framework, we aim to simultaneously balance strictly competing objectives, namely, flight safety, traffic flow, and communication reliability.

{
The H-MO-POMDP is formally defined by the tuple $\langle \mathcal{S}, \mathcal{A}, \mathcal{P}, \vec{\mathcal{R}}, \Omega, \mathcal{O}, \gamma_d \rangle$, where:
\begin{itemize}
    \item $\mathcal{S}$ is the global state space of the multi-UAV environment.
    \item $\mathcal{A}$ is the joint hybrid action space for hierarchical motion and communication.
    \item $\mathcal{P}(\mathcal{S}_{t+1} | \mathcal{S}_t, \mathbf{a}_t)$ represents the transition probability function, governed by the rigid-body physics engine and wireless channel stochasticity.
    \item $\vec{\mathcal{R}}$ is the multi-objective reward vector.
    \item $\Omega$ is the joint localized observation space available to the agents.
    \item $\mathcal{O}(\mathbf{o}_t | \mathcal{S}_t, \mathbf{a}_{t-1})$ is the joint observation probability function generating the partial states.
    \item $\gamma_d \in [0, 1)$ is the temporal discount factor. 
\end{itemize}
}

To effectively solve this complex H-MO-POMDP, decision-making authority over the tuple is distributed across two cognitive tiers, i.e., a slow-timescale HAPS meta-controller for global strategic orchestration and fast-timescale UAV edge-agents for localized tactical execution.

\begin{figure*}[ht]
    \centering
    \includegraphics[width=0.75\linewidth]{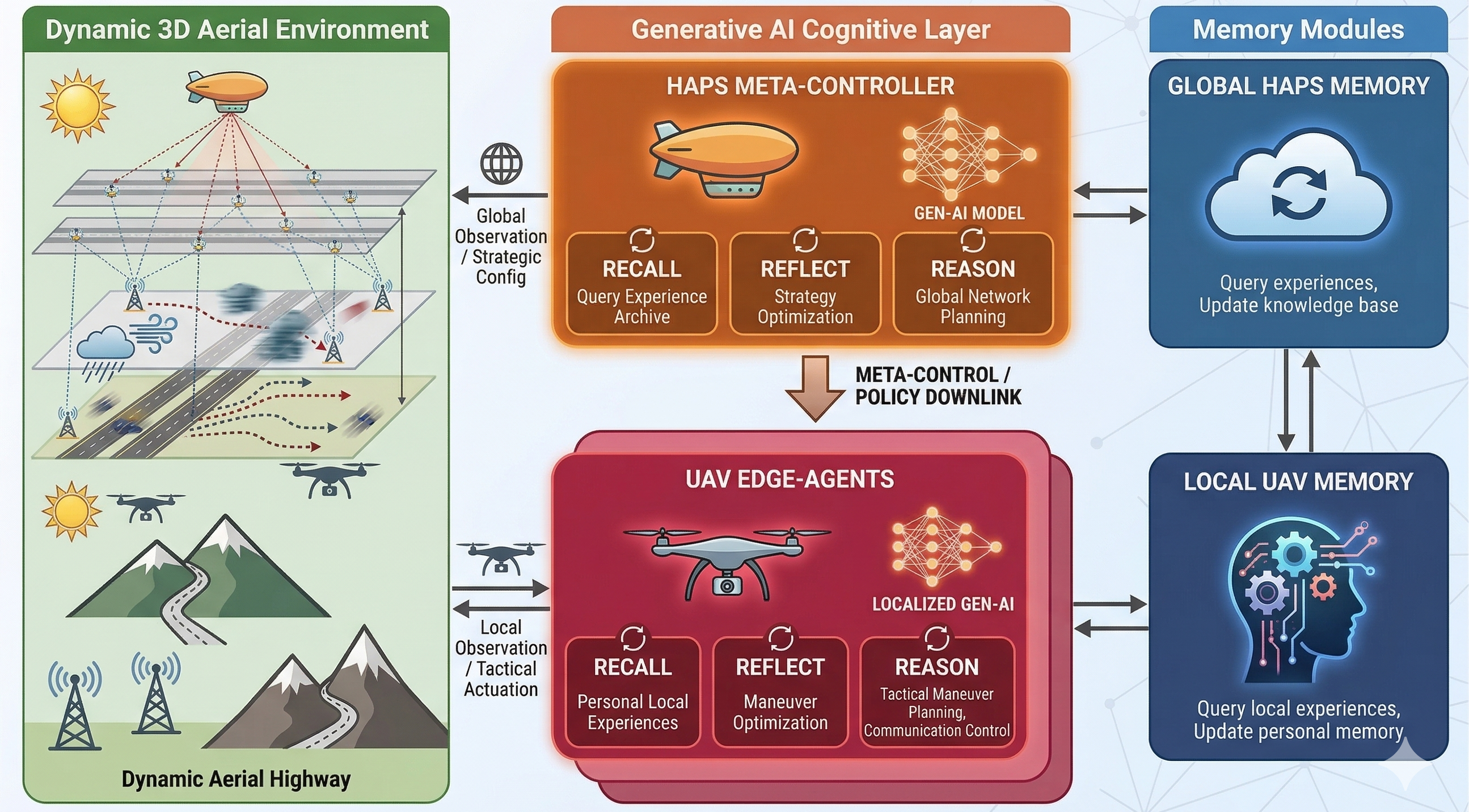}
    \caption{The generative AI-driven hierarchical control paradigm for multi-UAV networks in an ITNTN environment.}
    \label{fig:haps_hierarchical_framework}
\end{figure*}

\subsection{State  and Observation  Spaces} %($\mathcal{S}$) ($\Omega$)

\subsubsection{Global Environment State}%($\mathcal{S}$)
The global state $\mathcal{S}_t \in \mathcal{S}$ represents the ground truth tracked by the simulation environment. It includes the complete kinematic data (i.e., 3D position, velocity, attitude, and rotor RPMs) of all $M$ UAVs, governed by the rigid-body dynamics in Section \ref{subsec:dynamics_model}, alongside the precise network load of all base stations $\mathcal{B}$ and the HAPS $H$.

\subsubsection{HAPS Meta-Observation} %($\mathbf{o}^{\text{HAPS}}_t$)
The HAPS operates as a strategic orchestrator. Its observation space aggregates network-level metrics as
\begin{equation}
    \mathbf{o}^{\text{HAPS}}_t = \Bigl[ \ell^{\text{HAPS}}_t, \, C^{\text{HAPS}}_{\text{rem}}, \, \{d^{m,H}_t\}_{m \in \mathcal{M}}  , \{\text{WR}^{m,H}_t\}_{m \in \mathcal{M}} \Bigr],
\end{equation}
where $\ell^{\text{HAPS}}_t$ is the current HAPS bandwidth load, $C^{\text{HAPS}}_{\text{rem}}$ is the remaining capacity, and $d^{m,H}_t (\text{resp. } \text{WR}^{m,H}_t$) represents the distance between UAV $m$ and HAPS (resp. the weighted data rate of each connected UAV) at state $\mathcal{S}_t$.

{\subsubsection{UAV Local Observation}
Due to sensing and communication limitations, each UAV $m$ receives a local observation generated by the observation function $\mathcal{O}(\mathbf{o}^m_t | \mathcal{S}_t, \mathbf{a}^m_{t-1})$.
% To accurately reflect the state outputs of the \textit{gym-pybullet-drones} environment alongside our network parameters, 
Inspired by \cite{panerati2021learning,yan2024generalized}, the local observation vector is defined by
\begin{equation}
    \mathbf{o}^m_t = \Bigl[ \tilde{\mathbf{x}}^m_t, \tilde{\mathbf{v}}^m_t, \tilde{\boldsymbol{\theta}}^m_t, \tilde{\boldsymbol{\omega}}^m_t, \tilde{\mathbf{P}}^m_t, \mathcal{N}_t^m, c^m_t, d_t^{m,c}, \text{SINR}_t^{m,c} \Bigr],
\end{equation}
where the tilde ($\sim$) denotes noisy onboard sensor readings. Specifically, $\tilde{\mathbf{x}}^m_t$ and $\tilde{\mathbf{v}}^m_t$ are the 3D position and linear velocity, $\tilde{\boldsymbol{\theta}}^m_t$ represents the attitude via Euler angles (roll, pitch, yaw), $\tilde{\boldsymbol{\omega}}^m_t$ is the angular velocity, and $\tilde{\mathbf{P}}^m_t$ captures the current RPMs of the four rotors. The term $\mathcal{N}_t^m$ encapsulates the relative positions and velocities of nearby neighbor UAVs within a designated perception radius. For the telecommunication state, $c^m_t \in \mathcal{B} \cup \{HAPS\} $ denotes the currently associated serving node, and $d_t^{m,c} (\text{resp. } \text{SINR}_t^{m,c}$) represents the corresponding distance (resp. real-time link quality) between the node $c^m_t$ and UAV $m$.}

\subsection{Hierarchical Action Space}
\label{subsec:action_spaces}
Because decision-making is distributed across two cognitive tiers, the global action space $\mathcal{A}$ is decoupled into strategic HAPS meta-actions and tactical UAV edge-actions.

\subsubsection{HAPS Meta-Action Space ($\mathcal{A}^{\text{HAPS}}$)}
Operating on the slow timescale $T_{\text{HAPS}}$, the HAPS meta-controller determines the vertical network topology to prevent capacity saturation. Its discrete action space is defined as:
\begin{equation}
    \mathbf{a}^{\text{HAPS}}_t \in \bigl\{ \texttt{Offload}(m), \, \texttt{Recall}(m), \, \texttt{Idle} \bigr\},
\end{equation}
where $\texttt{Offload}(m)$ forces a specific UAV $m$ to disconnect from the HAPS and associate with a terrestrial BS, $\texttt{Recall}(m)$ allows a previously offloaded UAV to re-establish a HAPS link, and $\texttt{Idle}$ maintains the current network associations.

\subsubsection{UAV Tactical Action Space ($\mathcal{A}^m$)}
Operating on the fast timescale $T_{\text{fast}}$ (where $T_{\mathrm{fast}}<T_{\mathrm{HAPS}}$), each UAV edge-agent $m$ selects a joint hybrid action $\mathbf{a}^m_t = [\mathbf{a}^m_{\text{mot}, t}, \mathbf{a}^m_{\text{tele}, t}]^\top$ to handle immediate motion and communication dynamics.

{\textbf{1) Continuous Motion Action:} To fully leverage the high-fidelity rigid body dynamics, the physical motion action specifies the continuous target rotational speeds for the four rotors:
\begin{equation}
    \mathbf{a}^m_{\text{mot}, t} = [P_{1,m,t}, P_{2,m,t}, P_{3,m,t}, P_{4,m,t}]^\top,
\end{equation}
where $P_{k,m,t} \in [P_{\min}, P_{\max}]$ represents the RPM for each rotor $k \in \{1, \dots, 4\}$. As previously established, rather than forcing the LLM to directly output these continuous variables, the semantic directives of the LLM are deterministically mapped into this continuous action vector via the onboard motion decision decoder.}

\textbf{2) Discrete Telecommunication Action:} Simultaneously, the UAV agent selects a localized network association command:
\begin{equation}
    \mathbf{a}^m_{\text{tele}, t} \in \bigl\{ \text{Stay}, \, \text{HO to } b \in \mathcal{B}, \, \text{Request HAPS} \bigr\}.
\end{equation}
It is important to note the hierarchical dependency where a UAV's tactical decision to $\text{Request HAPS}$ is ultimately {governed} by the strategic $\mathbf{a}^{\text{HAPS}}_t$ decisions made by the HAPS meta-controller.

\subsection{Vectorized Multi-Objective Reward}

To address the H-MO-POMDP, we define the reward as a vector comprising distinct and often conflicting objectives. For each UAV $m \in \mathcal{M}$ at timestep $t$, the vectorized reward $\vec{\mathcal{R}}^m_t$ is formulated as
\begin{equation}
\label{eq:multi-object-rewards}
\vec{\mathcal{R}}^m_t=
\begin{bmatrix}
R_{\text{tran},t}^m &
R_{\text{tele},t}^m &
- C_{\text{safe},t}^m &
- C_{\text{HO},t}^m
\end{bmatrix}^\top,
\end{equation}
where the components are mathematically defined to balance physical navigation efficiency and network reliability as follows:

\begin{itemize}
    \item \textbf{Transportation Efficiency ($R_{\text{tran},t}^m$):} This metric rewards forward progress while penalizing excessive energy consumption and unstable flight. It is modeled as
\begin{equation}
\begin{aligned}
R_{\text{tran},t}^m 
&= \alpha_1 \exp\big(-\| \mathbf{x}^m_t - {\mathbf{x}_{\text{target}}^m} \|_2\big) \\
&\quad - \alpha_2 \| \mathbf{a}^m_{\text{mot}, t} \|_2^2 
      - \alpha_3 \| \boldsymbol{\theta}^m_t \|_2^2,
\end{aligned}
\end{equation}
    {where the first term incentivizes reducing the 3D Euclidean distance to its specific target waypoint ($\mathbf{x}_{\text{target}}^m$)}, the second term penalizes aggressive changes in motor RPMs in terms of exceeding energy consumption, and the third term penalizes extreme Euler angles ($\boldsymbol{\theta}^m_t$) to maintain aerodynamic stability.

    \item \textbf{Telecommunication ($R_{\text{tele},t}^m$):} It rewards the achievable data rate based on the selected serving node, scaled by the network capacity constraints. It is written as
    \begin{equation}
        R_{\text{tele},t}^m = \text{WR}_t^{m,c},
    \end{equation}
    ensuring the agent favors high-SINR links without overloading the HAPS or {BSs}.
    
    \item \textbf{Safety Penalty ($C_{\text{safe},t}^m$):} A sparse {yet} severe penalty is triggered to prevent collisions in the dense aerial highway. It is expressed by
    \begin{equation}
        C_{\text{safe},t}^m = 
        \begin{cases} 
            \rho_{\text{crash}}, & \text{if } \min_{j \in \mathcal{N}_t^m} \| \mathbf{x}^m_t - \mathbf{x}^j_t \|_2 < d_{\text{safe}} \\
            0, & \text{otherwise},
        \end{cases}
    \end{equation}
    where $d_{\text{safe}}$ is the minimum allowable physical separation distance between UAVs.
    
    \item \textbf{Handover Cost ($C_{\text{HO},t}^m$):} It is a dense penalty applied when the discrete telecommunication action $\mathbf{a}^m_{\text{tele}, t}$ triggers a network switch, reflecting the signaling overhead and transient latency as
    \begin{equation}
        C_{\text{HO},t}^m = \beta \, \mathbb{I}(c^m_t \neq c^m_{t-1}),
    \end{equation}
    where $\mathbb{I}(\cdot)$ is the indicator function and $\beta$ is the handover severity weight.
\end{itemize}

During execution, our generative AI-driven hierarchical LLM architecture navigates this adversarial multi-objective space. The semantic reasoning of the two-tier LLMs evaluates the tradeoff between physical velocity ($R_{\text{tran}}$) and link stability ($C_{\text{HO}}$).

% \subsection{HAPS Meta-Controller for Strategic Network Management}
\label{subsec:haps_realization}

\begin{figure*}[ht]
    \centering
    \includegraphics[width=\linewidth]{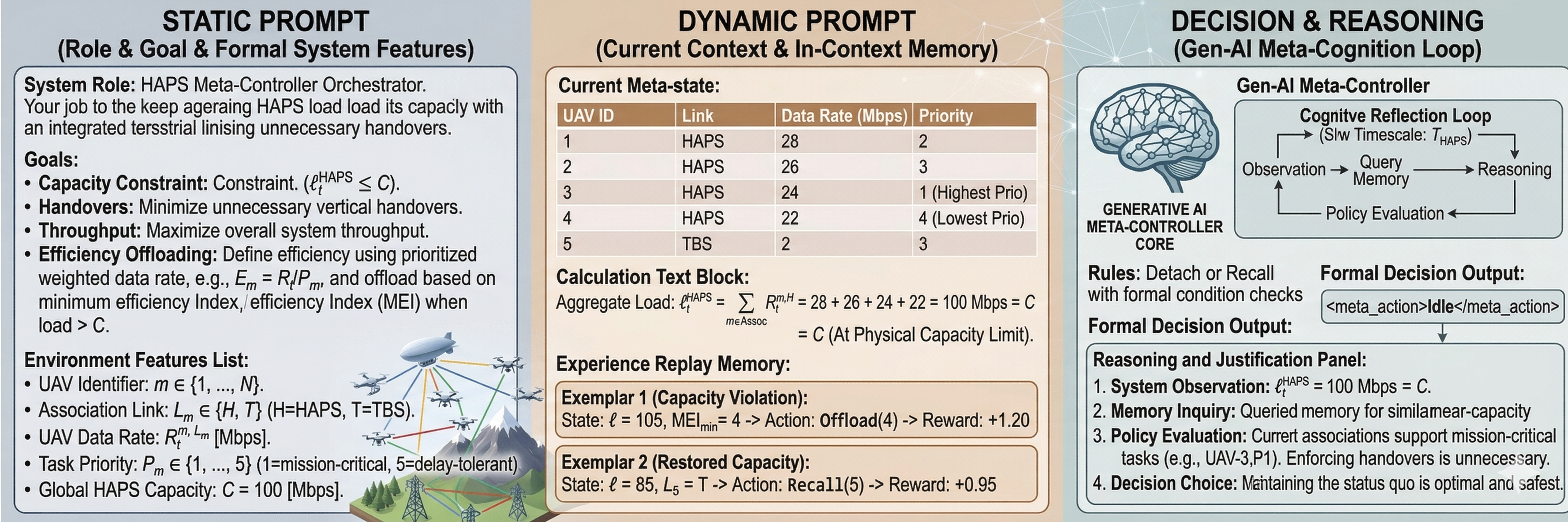}
    \caption{LLM-based HAPS Meta-controller Decision Process, illustrating the synthesis of the static role, dynamic observation state, and retrieved history memories into a structured meta-action.}
    \label{fig:haps-prompt}
\end{figure*}

\section{Collaborative LLM-Driven Cognitive Framework}
\label{sec:algorithm}

To effectively navigate the highly dynamic ITNTN environment and solve the H-MO-POMDP defined in Section \ref{sec:mdp_formulation}, we detail here the operational realization of the proposed collaborative LLM-driven cognitive framework. In contrast to conventional DRL pipelines that rely strictly on parameterized value updates, the proposed architecture (illustrated in Fig.~\ref{fig:haps_hierarchical_framework}) utilizes prompt-conditioned reasoning, memory retrieval, and structured action decoding. The system follows a closed-loop cognitive paradigm as follows: {observe $\rightarrow$ retrieve $\rightarrow$ reason $\rightarrow$ act $\rightarrow$ reflect $\rightarrow$ update memory}.

\subsection{Overall Multi-Timescale Execution}
\label{subsec:multitimescale_execution}

To prevent the latency bottleneck inherent in generative AI inference, the framework operates on a dual-timescale coordinated execution model, specifically:
\begin{itemize}
    \item \textbf{HAPS Meta-Tier (Slow Timescale, $T_{\mathrm{HAPS}}$):} The HAPS meta-controller executes globally, regulating load distribution, vertical handovers, and congestion prevention.
    \item \textbf{UAV Tactical Tier (Fast Timescale, $T_{\mathrm{fast}}$):} Each UAV edge-agent determines localized continuous motion and discrete communication actions based on real-time channel fluctuations and current network guidance.
\end{itemize}
This bi-level execution drastically reduces the complexity of full joint control while preserving coordination between global network management and localized UAV adaptation.

\subsection{Task Discretization and Prompt Construction}
\label{subsec:prompt_construction}

Because LLMs inherently process semantic tokens rather than continuous numerical tensors, numerical observations must be converted via a task discretization function $\mathcal{T}(\cdot)$. For a given tier-specific observation (either the global HAPS meta-state $\mathbf{o}^{\mathrm{HAPS}}_{t}$ or the local UAV state $\mathbf{o}^{m}_{t}$), the dynamic state prompt is generated as $\mathbf{p}_{t}^{\mathrm{dyn}} = \mathcal{T}(\mathbf{o}_{t})$. This function maps numerical bounds into categorical text.
% (e.g., mapping $d_t^{m,c}$ to ``Strong Signal'' or relative neighbor distances to ``Imminent Collision Risk'').

Both the HAPS controller and UAV agents generate decisions through this structured prompt-based reasoning. As illustrated in Fig. \ref{fig:haps-prompt}, at each decision epoch, the complete prompt $\mathcal{P}_t$ is synthesized as
\begin{equation}
    \mathcal{P}_t = \mathcal{P}^{\mathrm{stat}} \oplus \mathbf{p}^{\mathrm{dyn}}_t \oplus \mathcal{P}^{\mathrm{mem}}_t,
\end{equation}
where $\oplus$ denotes string concatenation, and the prompt components are defined by
\begin{itemize}
    \item \textbf{Static Prompt ($\mathcal{P}^{\mathrm{stat}}$):} It defines the agent's role, task objectives, action constraints, and required output format. For the HAPS controller, it enforces the goal of maintaining the load below $C_{\max}^{\mathrm{HAPS}}$ while minimizing HO events. For the UAV agents, it enforces localized 3D flight safety, trajectory goals, and interference constraints.
    \item \textbf{Dynamic Prompt ($\mathbf{p}^{\mathrm{dyn}}_t$):} It describes the real-time discretized observable state. The HAPS dynamic prompt includes the current association profile, weighted rates $\mathrm{WR}_t^{m,c}$, and aggregate load $\ell_t^{\mathrm{HAPS}}$. The UAV dynamic prompt details its kinematics, neighbor states $\mathcal{N}_t^m$, serving node, and localized SINR.
    \item \textbf{Memory Prompt ($\mathcal{P}^{\mathrm{mem}}_t$):} It contains the top-$K$ historical examples retrieved from the episodic memory pool. These examples act as in-context demonstrations, guiding the LLM toward high-quality decisions under similar environmental conditions without gradient updates.
\end{itemize}

% \textbf{Motivation \& Mission:} 
In a dense ITNTN, individual UAVs lack global network visibility. Greedy and uncoordinated BS associations inevitably lead to severe local terrestrial congestion and resource starvation. The HAPS meta-controller operates as a strategic orchestrator to resolve this. Indeed, it keeps the aggregate HAPS load strictly below the capacity limit ($\ell_t^{\mathrm{HAPS}} \leq C_{\max}^{\mathrm{HAPS}}$) while maximizing overall network throughput and minimizing unnecessary vertical HOs.

% \textbf{Operational Execution:}
As illustrated in Fig.~\ref{fig:haps-prompt}, at each slow-timescale decision epoch, the HAPS controller synthesizes its structured prompt $\mathcal{P}^{\mathrm{HAPS}}_t$. Operating purely as a generative AI orchestrator, the meta-action is sampled from the LLM policy $\pi_{\mathrm{HAPS}}$:
% \begin{equation}\label{eq-haps-llm}
%     \mathbf{a}_t^{\mathrm{HAPS}} \sim \pi_{\mathrm{HAPS}}\bigl(\cdot \mid \mathcal{P}^{\mathrm{HAPS}}_t\bigr) \in \{\texttt{Offload}(\mathcal{U}), \texttt{Recall}(\mathcal{U}), \texttt{Idle}\},
% \end{equation}
\begin{equation}\label{eq-haps-llm}
\begin{aligned}
\mathbf{a}_t^{\mathrm{HAPS}} 
&\sim \pi_{\mathrm{HAPS}}\bigl(\cdot \mid \mathcal{P}^{\mathrm{HAPS}}_t\bigr) \\
&\in \{\texttt{Offload}(\mathcal{U}), \texttt{Recall}(\mathcal{U}), \texttt{Idle}\},
\end{aligned}
\end{equation}
where $\mathcal{U} \subseteq \mathcal{M}$ represents the specific subset of UAVs selected for network reconfiguration. The chosen action is immediately broadcast to the affected UAVs, and the global association matrix is updated before the lower-tier edge-agents execute their tactical decisions.

To continuously evaluate the quality of these strategic decisions, the HAPS computes a global system reward, defined as
% \begin{equation}
%     R_t^{\mathrm{HAPS}} = \eta_1 \sum_{m \in \mathcal{M}} \mathrm{WR}_t^{m,c} - \eta_2 \mathbb{I}\bigl(\ell_t^{\mathrm{HAPS}} > C_{\max}^{\mathrm{HAPS}}\bigr) - \eta_3 \sum_{m \in \mathcal{M}} \mathbb{I}(\mathrm{HO}_t^m),
% \label{eq:haps_reward_final}
% \end{equation}
\begin{equation}\label{eq:haps_reward_final}
\begin{aligned}
R_t^{\mathrm{HAPS}} 
&= \eta_1 \sum_{m \in \mathcal{M}} \mathrm{WR}_t^{m,c} - \eta_2 \mathbb{I}\bigl(\ell_t^{\mathrm{HAPS}} > C_{\max}^{\mathrm{HAPS}}\bigr)\\ 
&- \eta_3 \sum_{m \in \mathcal{M}} \mathbb{I}(\mathrm{HO}_t^m),
\end{aligned}
\end{equation}
where $\eta_1, \eta_2, \eta_3$ are strategic balancing weights. This reward function is used exclusively for experience evaluation and memory updates, rather than for gradient-based parameter backpropagation. The complete strategic execution loop is summarized in Algo.~\ref{alg:meta_controller}.

\begin{algorithm}[ht]
\caption{LLM-driven HAPS Meta-Controller for Network Reconfiguration}
\label{alg:meta_controller}
\SetKwInOut{Input}{Input}
\SetKwInOut{Output}{Output}

\Input{Initial meta-observation $\mathbf{o}^{\text{HAPS}}_0$, UAV set $\mathcal{M}$, HAPS capacity $C^{\text{HAPS}}_{\max}$}
\Output{Strategic UAV association directives $\mathbf{a}^{\text{HAPS}}_t$}
Initialize LLM meta-controller parameters and global memory buffer\;
\For{each timestep $t$ in episode}{
    \tcp{Execute only on the slow HAPS timescale}
    \If{$t \pmod{T_{\text{HAPS}}} == 0$}{
        Observe meta-observation $\mathbf{o}^{\text{HAPS}}_t$ (load $\ell^{\text{HAPS}}_t$, data rates $\text{WR}^{m,H}_t$)\;
        Identify current HAPS-associated UAV subset $\mathcal{M}_H \subseteq \mathcal{M}$\;
        
        \uIf{$\ell^{\text{HAPS}}_t > C^{\text{HAPS}}_{\max}$}{
            Identify subset $\mathcal{U} \subseteq \mathcal{M}_H$ with lowest data rates or least critical tasks\;
            Action $\mathbf{a}^{\text{HAPS}}_t \leftarrow$ \texttt{Offload($\mathcal{U}$)} to terrestrial BS\;
        }
        \uElseIf{$\ell^{\text{HAPS}}_t < C^{\text{HAPS}}_{\max}$ and offloaded UAVs exist}{
            Identify eligible subset $\mathcal{U}' \notin \mathcal{M}_H$ for reattachment\;
            Action $\mathbf{a}^{\text{HAPS}}_t \leftarrow$ \texttt{Recall($\mathcal{U}'$)} to HAPS\;
        }
        \Else{
            Action $\mathbf{a}^{\text{HAPS}}_t \leftarrow$ \texttt{Idle}\;
        }
        Broadcast action $\mathbf{a}^{\text{HAPS}}_t$ to the respective UAV edge-agents\;
        
        \tcp{Observe feedback and update global memory}
        Observe next meta-state $\mathbf{o}^{\text{HAPS}}_{t+1}$\;
        Compute global system reward $R^{\text{HAPS}}_t$ via Eq. (\ref{eq:haps_reward_final})\;
        Update meta-policy $\pi_{\text{HAPS}}$ via LLM cognitive reflection on $(\mathbf{o}^{\text{HAPS}}_t, \mathbf{a}^{\text{HAPS}}_t, R^{\text{HAPS}}_t, \mathbf{o}^{\text{HAPS}}_{t+1})$\;
    }
}
\end{algorithm}

\subsection{UAV Edge-Agent for Tactical Joint Control}
\label{subsec:uav_realization}

\begin{figure*}[t]
    \centering
    \includegraphics[width=0.75\linewidth]{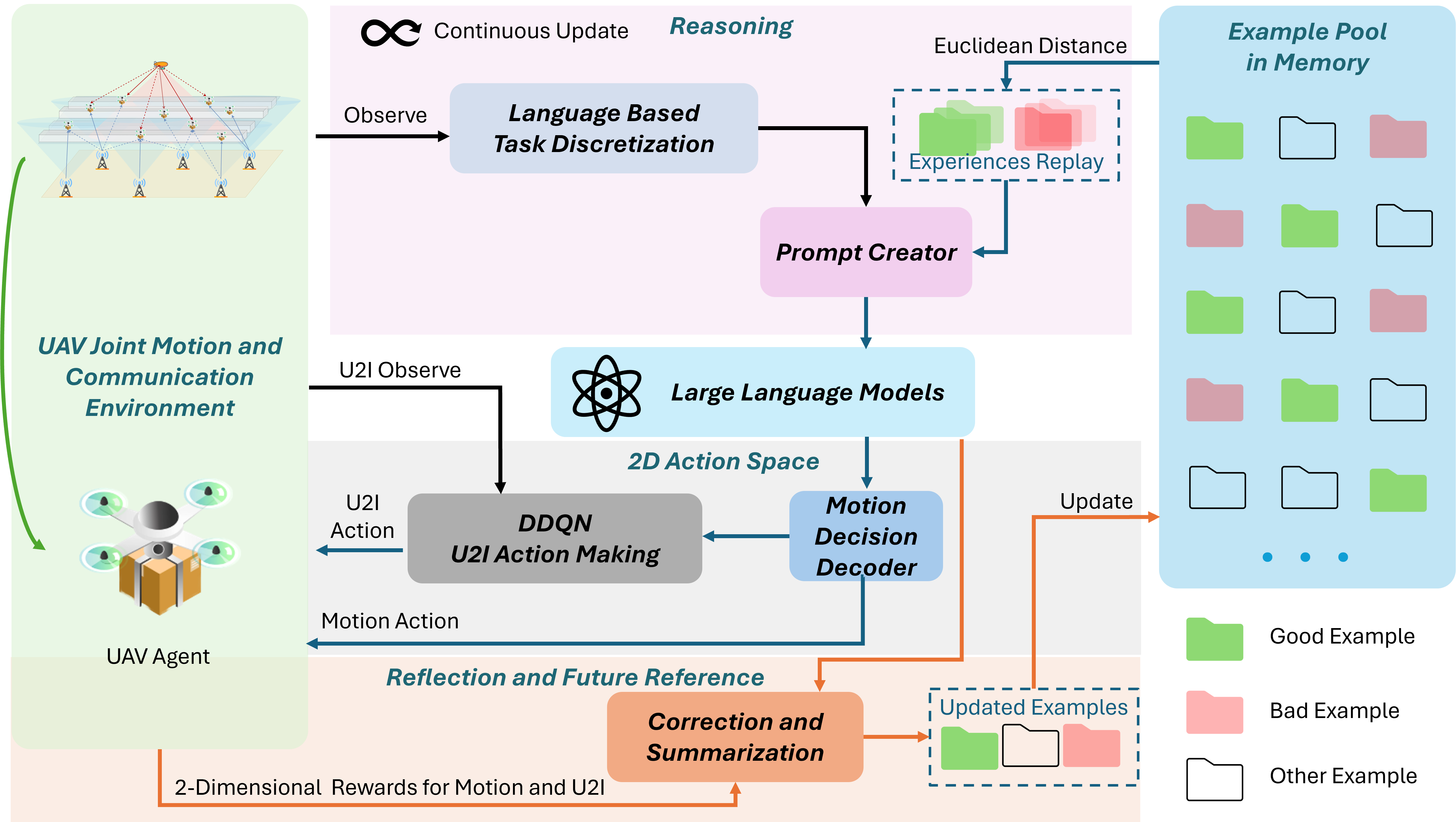}
    \caption{The architectural framework of the proposed UAV edge-agent.}
    \label{fig:uav_edge_agent}
\end{figure*}

% \textbf{Motivation \& Mission:} 
At the edge, UAVs navigating the 3D aerial highway face highly dynamic physical threats like collisions and rapid RF channel fading that cannot wait for slow-timescale interventions from the HAPS. The UAV edge-agent acts as a localized tactical controller. It leverages the advanced spatial reasoning of an onboard LLM to ensure collision-free physical trajectories, while simultaneously utilizing a high-frequency DRL policy to maintain an optimal communication link.

As depicted in Fig.~\ref{fig:uav_edge_agent}, the UAV agent strictly decouples its action space to satisfy these real-time latency constraints, as follows:

\textbf{1) Continuous Motion Control:} The onboard LLM evaluates the local prompt $\mathcal{P}^{m}_t$ to generate a high-level spatial directive $\mathbf{a}^{\text{sem}, m}_t \sim \pi_{\mathrm{UAV}}(\cdot \mid \mathcal{P}^{m}_t)$. A deterministic Motion Decision Decoder $\mathcal{G}(\cdot)$ then translates this semantic intent into the continuous rotor RPMs required by the rigid-body physics engine, i.e., 
\begin{equation}\label{eq-uav-mot}
    \mathbf{a}^{m}_{\mathrm{mot},t} = \mathcal{G} \bigl( \mathbf{a}^{\text{sem}, m}_t \bigr),
\end{equation}
yielding the precise $[P_{1,m,t}, P_{2,m,t}, P_{3,m,t}, P_{4,m,t}]^\top$ motor thrusts.

\textbf{2) Discrete Telecommunication Control:} Concurrently, because sub-millisecond RF channel fluctuations cannot wait for slow LLM inference cycles, the discrete telecommunication handover is processed by a parallel DDQN parameterized by $\theta$. The DDQN selects the optimal action based on the current local observation and the decided motion intent as
\begin{equation}\label{eq-uav-tele}
    \mathbf{a}^{m}_{\mathrm{tele},t} = \arg\max_{\mathbf{a}} Q\bigl( \mathbf{o}^{m}_{t}, \mathbf{a}^{m}_{\mathrm{mot},t}, \mathbf{a}; \theta \bigr).
\end{equation}
This hybrid architecture allows the UAV to execute the joint action $\mathbf{a}_t^m = [\mathbf{a}^{m}_{\mathrm{mot},t}, \mathbf{a}^{m}_{\mathrm{tele},t}]^\top$ seamlessly. The complete tactical loop for the edge-agent is summarized in Algo.~\ref{alg:edge_cognitive_loop}.
For the semantic reasoning layer, this mechanism facilitates a lightweight, self-improving cognitive loop without the severe computational overhead of gradient-based parameter backpropagation, leaving intensive weight updates to the DDQN.

\begin{algorithm}[t]
\caption{UAV Edge-Agent Tactical Cognitive Loop}
\label{alg:edge_cognitive_loop}
\SetKwInOut{Input}{Input}
\SetKwInOut{Output}{Output}

\Input{Initial observation $\mathbf{o}^m_0$, Memory pool $\mathcal{D}^m$, DDQN parameters $\theta$, Threshold $\rho_{\text{thresh}}$, Weights $\vec{\mathbf{w}}$}
\For{each timestep $t$ in episode}{
    Observe continuous local state $\mathbf{o}^m_t$\;

    \If{$t \pmod{T_{\text{fast}}} == 0$}{
        \tcp{Slow Timescale: LLM Spatial Reasoning}
        \If{$t \pmod{T_{\mathrm{HAPS}}} == 0$ \textnormal{\textbf{or}} $t \pmod{T_{\text{LLM}}} == 0$}{
            Discretize state: $\mathbf{p}^{\text{dyn}}_t \leftarrow \mathcal{T}(\mathbf{o}^m_t)$\;
            Retrieve top-$K$ memories $\mathcal{D}^m_{\text{top-}K}$ using Euclidean distance $d_i$\;
            Generate prompt $\mathcal{P}^{m}_t = \mathcal{P}^{\mathrm{stat}} \oplus \mathbf{p}^{\text{dyn}}_t \oplus \mathcal{P}^{\mathrm{mem}}_t$\;
            Generate semantic directive: $\mathbf{a}^{\text{sem}, m}_t \sim \pi_{\mathrm{UAV}}(\cdot \mid \mathcal{P}^{m}_t)$\;
        }
        
        \tcp{Fast Timescale: Execution \& U2I}
        Decode motion action (Eq. \ref{eq-uav-mot}): $\mathbf{a}^m_{\text{mot}, t} = \mathcal{G}(\mathbf{a}^{\text{sem}, m}_t)$\;
        Select U2I action via DDQN (Eq. \ref{eq-uav-tele}): $\mathbf{a}^m_{\text{tele}, t} = \arg\max_{\mathbf{a}} Q(\mathbf{o}^m_t, \mathbf{a}^m_{\text{mot}, t}, \mathbf{a}; \theta)$\;
        Execute joint 2D action $\mathbf{a}_t^m = [\mathbf{a}^{m}_{\text{mot},t}, \mathbf{a}^{m}_{\text{tele},t}]^\top$\;
        Observe multi-objective reward $\vec{\mathcal{R}}^m_t$ and next state $\mathbf{o}^m_{t+1}$\;
        
        \tcp{Cognitive Reflection and Memory Update}
        \eIf{$\vec{\mathbf{w}}^\top \vec{\mathcal{R}}^m_t < \rho_{\mathrm{thresh}}$}{
            Trigger Reflection: Prompt LLM with $(\mathbf{p}^{\text{dyn}}_t, \mathbf{a}^m_t, \vec{\mathcal{R}}^m_t)$\;
            Generate linguistic correction summary $c_t^m$\;
        }{
            Set $c_t^m \leftarrow \varnothing$ 
            % \tcp*{Store as positive exemplar}
        }
        {Update memory: $\mathcal{D}^m \leftarrow \mathcal{D}^m \cup \{(\mathbf{o}^m_t, \mathbf{a}^m_t, \vec{\mathcal{R}}^m_t, \mathbf{o}^m_{t+1}, c_t^m) \}$\;}
        Update localized DDQN parameters $\theta$ using Bellman error\;
    }
}
\end{algorithm}

\subsection{Cognitive Reflection and Memory Retrieval}
\label{subsec:memory_retrieval}

To improve zero-shot robustness and adaptability in highly dynamic ITNTN environments, both the HAPS meta-controller and the UAV edge-agents maintain independent episodic memory buffers. The buffer for each cognitive tier is defined as
\begin{equation}
    \mathcal{D} = \{(\mathbf{o}_i, \mathbf{a}_i, \mathcal{R}_i, \mathbf{o}'_i, \mathcal{C}_i)\}_{i=1}^{N},
\end{equation}
where $\mathbf{o}_i$ is the localized observation ($\mathbf{o}^{\mathrm{HAPS}}_t$ or $\mathbf{o}^m_t$), $\mathbf{a}_i$ is the executed action, $\mathcal{R}_i$ is the tier-specific reward ($R^{\mathrm{HAPS}}_t$ or the multi-objective vector $\vec{\mathcal{R}}^m_t$), $\mathbf{o}'_i$ is the subsequent state, and \textcolor{blue}{$\mathcal{C}_i$} is a textual correction summary generated during past reflections.

%{\color{purple} Zijiang comment: Now, $\vec{\mathcal{R}}^m_t$  is defined in Eq. \ref{eq:multi-object-rewards} }

At decision epoch $t$, the current observation $\mathbf{o}_t$ is compared against stored historical states to identify highly relevant contextual precedents. We employ a Euclidean distance similarity measure applied to the normalized state embeddings $\phi(\cdot)$, i.e.,
\begin{equation}
    d_i = \|\phi(\mathbf{o}_t)-\phi(\mathbf{o}_i)\|_2.
\end{equation}
The top-$K$ experiences with the smallest distance $d_i$ are retrieved to construct the memory prompt $\mathcal{P}^{\mathrm{mem}}_t$. 

Following the execution of an action, the environment returns the reward signal and the next state. If the decision yields favorable performance, the tuple is stored directly as a positive exemplar ($c_t = \varnothing$). However, if the reward falls below a safety threshold, mathematically defined for the UAV $m$ as $\vec{\mathbf{w}}^\top \vec{\mathcal{R}}^m_t < \rho_{\text{thresh}}$ (indicating a near-miss collision or an inefficient handover loop), a cognitive reflection step is triggered. The LLM deduces the causality of the suboptimal action, generates a linguistic correction note $c_t$, and appends the updated experience to $\mathcal{D}$. This mechanism facilitates a lightweight, self-improving cognitive loop without the severe computational overhead of gradient backpropagation.

\section{Numerical Results and Discussion}
\label{sec:numerical_results}

To rigorously evaluate the proposed generative AI-driven hierarchical framework, we conduct extensive experiments focusing on both the qualitative cognitive reasoning of the LLM agents and the quantitative system-level performance under varying network densities.

\subsection{Simulation Setup and Parameters}
\label{subsec:sim_setup}

%\textbf{1) High-Fidelity 3D Environment:}
We adopt a high-fidelity 3D physics environment built upon  \cite{panerati2021learning}. This engine numerically integrates the rigid-body dynamics detailed in Section~\ref{subsec:dynamics_model}, capturing precise multi-rotor thrusts, torques, drag, and ground effects at a simulation frequency of 20~Hz ($\Delta t = 0.05$~s). The UAVs are modeled with a mass of $M_u = 1.5$~kg, representing standard commercial quadrotors capable of outdoor flight, and the minimum physical safety separation to avoid collision penalties is defined as $d_{\text{safe}} = 5.0$~m. The aerial highway is modeled as a $1000 \times 1000 \times 300$~m$^3$ airspace containing static cylindrical no-fly zones. To evaluate scalability, the number of active UAVs varies across $M \in \{5, 10, 15, 20, 25, 30\}$.

%\textbf{2) ITNTN Infrastructure:}
The communication network consists of one HAPS meta-node operating at an altitude of 20~km and $B = 4$ TBSs distributed uniformly across the ground plane. The HAPS operates at a carrier frequency of 2.0~GHz with a total allocated bandwidth of $B_{\mathrm{HAPS}} = 20$~MHz and a strict aggregated capacity limit of $C^{\text{HAPS}}_{\max} = 100$~Mbps. The terrestrial BSs operate at 2.1~GHz with a transmit power of $P_T = 40$~dBm and a peak antenna gain of $B_{\max} = 8$~dBi, utilizing 3GPP 3D antenna radiation patterns to compute accurate G2A line-of-sight probabilities and SINR. For the UAV uplink transmissions, the maximum transmit power is capped at $P_{\max} = 23$~dBm. The environment assumes a noise power spectral density of $N_0 = -174$~dBm/Hz, and the UAV-HAPS channels are modeled with a Rician LoS factor of $K = 15$~dB. Each serving node maintains a strict capacity quota of $Q_c = 5$ concurrent users.

%\textbf{3) LLM Deployment, Cognitive Agents, and Reward Setup:}
To simulate a realistic edge-cloud architecture, the LLM reasoning modules are deployed using the \textsc{Ollama} inference framework \cite{ollama2024} on a high-performance computing cluster. To ensure architectural consistency, we employ the state-of-the-art Qwen~3.5 model family \cite{yang2025qwen3}. The localized UAV edge-agents utilize the highly efficient \texttt{Qwen3.5-9B} model to ensure rapid tactical prompt processing within stringent edge-hardware constraints. Conversely, the global HAPS meta-controller utilizes the massive \texttt{Qwen3.5-122B} model, leveraging the cluster's A100 GPUs for complex, multi-agent strategic load balancing. 
%
% To satisfy the multi-rate execution constraint, the UAV LLM reasoning timescale is set to $T_{\text{LLM}} = 1.0$~s, whereas the DDQN and motion decoder execute at the fast physics timescale $T_{\text{fast}} = 0.05$~s. The global HAPS meta-controller regulates the network at a slower strategic timescale of $T_{\mathrm{HAPS}} = 5.0$~s. 
{
To satisfy the multi-rate execution constraint, the fundamental discrete time slot $t$ of the H-MO-POMDP formulation is strictly mapped to the fast physics execution step, such that one time slot equals $\Delta t = T_{\text{fast}} = 0.05$~s. Consequently, the upper-tier cognitive modules operate at integer multiples of this base time slot. The localized DDQN and motion decoder execute continuously at every time slot $t$. Conversely, the localized UAV LLM reasoning timescale is set to $T_{\text{LLM}} = 1.0$~s (executing every 20 time slots), and the global HAPS meta-controller regulates the network at a slower strategic timescale of $T_{\mathrm{HAPS}} = 5.0$~s (executing every 100 time slots).
}
Finally, for the H-MO-POMDP reward evaluation and cognitive reflection, the multi-objective safety threshold that triggers the LLM reflection module is set to $\rho_{\text{thresh}} = -10.0$. The severe collision penalty is defined as $\rho_{\text{crash}} = -100.0$, and the scalarization weights balancing the HAPS and UAV reward vectors are configured as $\eta_{\{1,2,3\}} = \{1.0, 50.0, 5.0\}$ and $\alpha_{\{1,2,3\}} = \{1.0, 0.1, 0.2\}$, respectively.
% Key simulation parameters are summarized in Table~\ref{tab:sim_parameters}.
% \textbf{4) Benchmark Algorithms:}
We compare our proposed \textit{Hierarchical LLM Framework} against the following state-of-the-art baselines:
\begin{itemize}
    \item \textbf{DDQN:} A standard Double Deep Q-Network agent tasked with managing both discrete flight waypoints and U2I handovers simultaneously, representing traditional single-objective DRL without semantic spatial reasoning.
    \item \textbf{Envelope-MORL \cite{yan2024generalized}:} A multi-objective RL approach that utilizes envelope updates to manage linear scalarization trade-offs between transportation efficiency and handover frequency.
\end{itemize}
\subsection{Case Studies of Hierarchical LLM Decision-Making}

\begin{figure*}[!ht]
    \centering
    \includegraphics[width=0.9\linewidth]{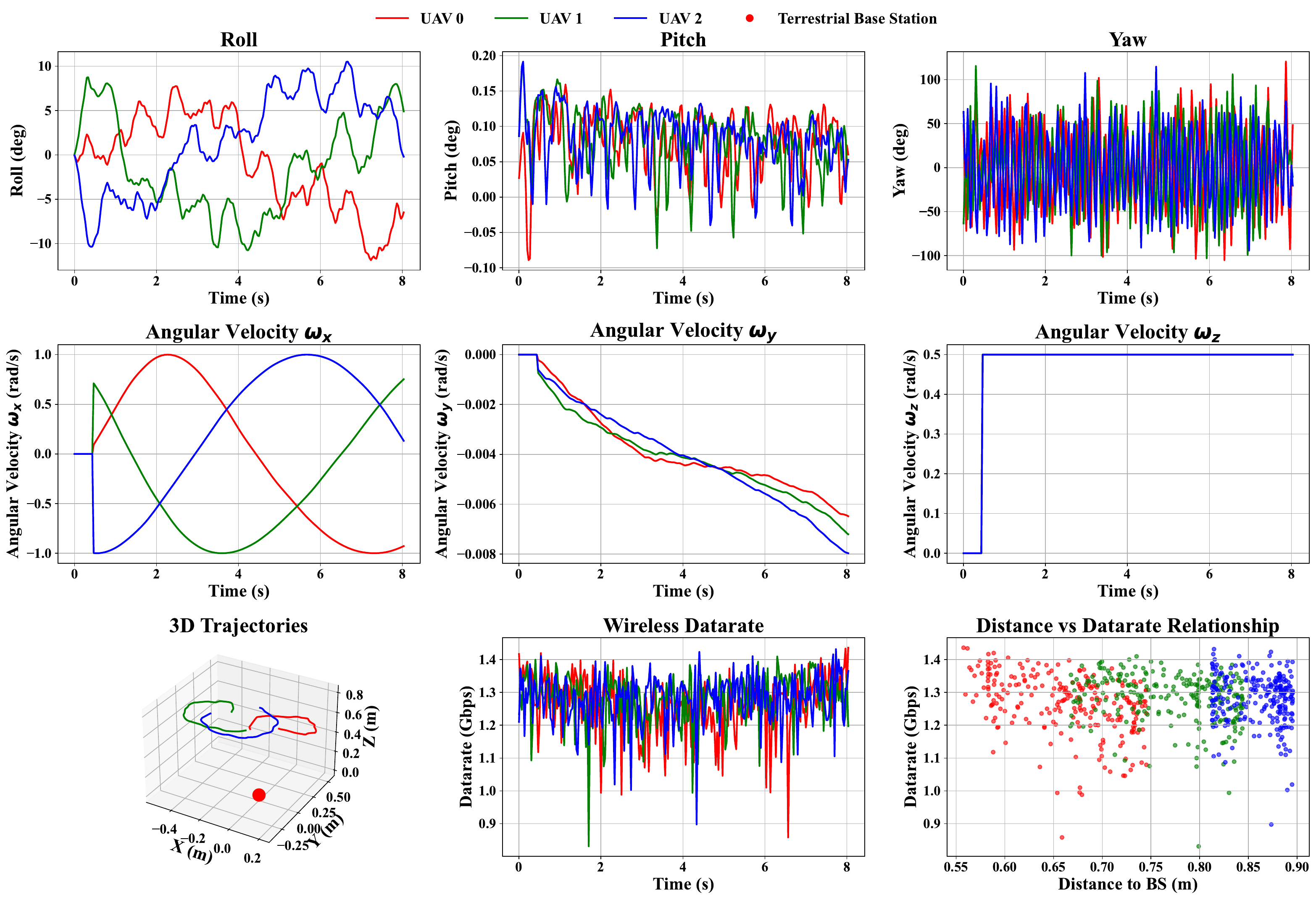}
    \caption{Comprehensive 20-second simulation profiles of two UAVs navigating an integrated terrestrial and non-terrestrial network (1 HAPS (not shown in the figure) , 1 TBS) using the proposed collaborative LLM-DDQN agent. }
    \label{fig:simulation-20s}
\end{figure*}

To intuitively validate the real-time cognitive reasoning and execution capabilities of our proposed framework, we extract a comprehensive 20-second flight episode. In this scenario, three UAVs navigate a shared 3D corridor while maintaining mission-critical data links within a network comprising one HAPS and one TBS. The operational state space is recorded and presented in Fig.~\ref{fig:simulation-20s}.

\begin{figure*}[ht]
    \centering
    \includegraphics[width=\linewidth]{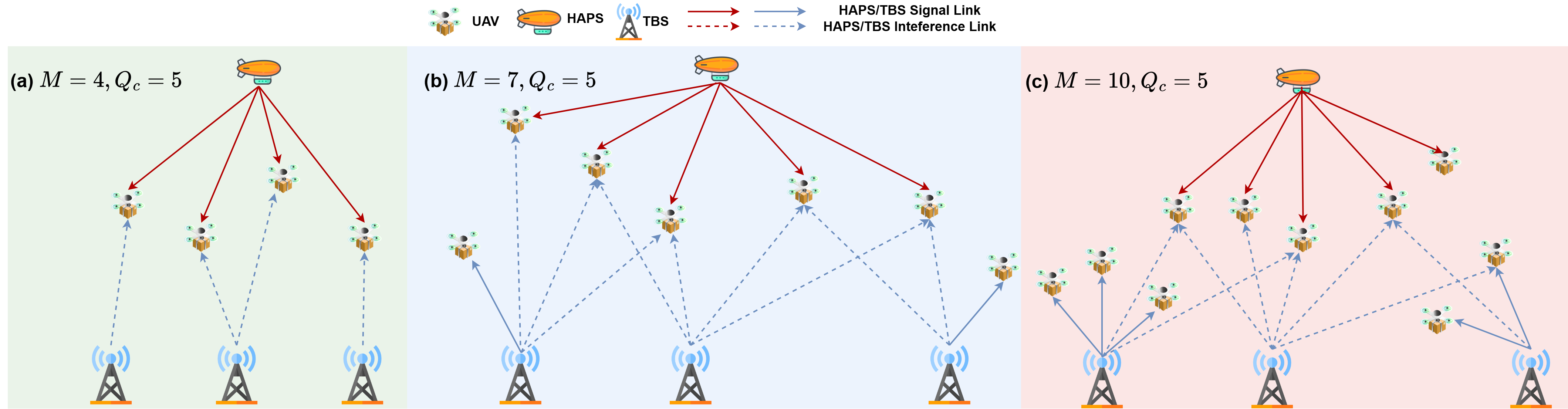}
    \caption{3D visualizations of the ITNTN network topology under varying traffic densities. (a) $M=4$: Demand is strictly below the capacity limit ($Q_c=5$). HAPS optimally serves all UAVs. (b) $M=7$: Demand exceeds capacity. The meta-controller actively offloads 2 UAVs to the TN to maintain HAPS stability. (c) $M=10$: Demand exceeds capacity. The meta-controller actively offloads 5 UAVs to the TN to maintain HAPS stability.}
    \label{fig:density_scenarios}
\end{figure*}
\textbf{1) Kinematic Stability and Spatial Reasoning:} 
A primary challenge in applying LLMs to autonomous flight is bridging the abstraction gap between discrete semantic commands and continuous, high-frequency aerodynamic stability. 
The attitude (roll, pitch) and angular velocity ($\boldsymbol{\omega}^m_t$) subplots (subfigures in {1st and 2nd} rows of Fig.~\ref{fig:simulation-20s}) demonstrate that the LLM's semantic directives are successfully decoded into stable physical maneuvers. The roll and pitch angles remain tightly bounded within safe operational thresholds of $\pm 15^\circ$, accompanied by smooth angular velocities near 0 rad/s (subfigures in 2nd row of Fig.~\ref{fig:simulation-20s}). This absolute smoothness definitively proves that the deterministic motion decoder successfully translates text commands into viable aerodynamics, completely avoiding the erratic shaking and violent overcorrections, compared to current existing approaches \cite{yan2023multi}.
As observed in the {3D Trajectories} subplot of Fig.~\ref{fig:simulation-20s} (subfigures in the last row), all three UAVs successfully advance toward their target zones while actively maintaining the strict $d_{\text{safe}}$ separation distance. Unlike traditional DRL agents that often artificially minimize safety penalties by adopting overly conservative hovering policies, the varying velocity profiles prove that the LLM's spatial discretization grants dynamic, zero-shot collision awareness. 

\textbf{2) Telecommunication Orchestration and Link Reliability:} 
Concurrently, the tactical tier must solve the H-MO-POMDP by ensuring that aggressive physical flight maneuvers do not {compromise} network connectivity. As depicted in the {Distance vs Datarate Relationship} {subplot of Fig. \ref{fig:simulation-20s}} (last row), the UAVs' physical distances vary drastically as they navigate the corridor, which naturally induces {substantial variations in} path loss and channel fading. 
However, despite these drastic spatial fluctuations, the {Average Datarate} remains remarkably stable, between 1.15 and 1.20~Gbps (subfigures in the last row of Fig. \ref{fig:simulation-20s}). This stability serves as the definitive proof of the proposed dual-timescale architecture, i.e., while the LLM governs the slow-timescale physical flight, the localized DDQN successfully executes rapid, high-frequency U2I handovers to compensate RF fading. By optimally timing these vertical handovers based on the LLM's semantic guidance, the dual-agent framework sustains the aggregate system data rate, intelligently relying on the terrestrial infrastructure's spatial multiplexing to handle the bulk of the traffic while reserving the HAPS's limited 100~Mbps capacity for critical coverage gaps.

\textbf{3) HAPS Meta-Controller Offloading Dynamics:} 
To explicitly validate the strategic load-balancing capabilities of the HAPS meta-controller, we simulate the aerial highway under varying traffic densities in Fig.~\ref{fig:density_scenarios}. The HAPS is constrained by a strict capacity limit, capable of concurrently serving a maximum quota of $Q_c = 5$ UAVs by intelligently offloading excess traffic to the terrestrial infrastructure.

In the low-density scenario (Fig.~\ref{fig:density_scenarios}a, $M=4$), the aggregate demand remains below the capacity threshold, and all active UAVs are optimally accommodated by the high-throughput HAPS. However, as traffic scales to medium density (Fig.~\ref{fig:density_scenarios}b, $M=7$), the network demand exceeds the non-terrestrial capacity. In this case, the meta-LLM anticipates this saturation as follows: Rather than allowing the non-terrestrial tier to degrade through interference and resource starvation, the meta-controller issues targeted \texttt{Offload($\mathcal{U}$)} directives, forcing two excess UAVs to execute vertical handovers to the terrestrial BSs. 

This load-balancing behavior scales robustly during the high-density surge (Fig.~\ref{fig:density_scenarios}c, $M=10$), where the meta-controller orchestrates a massive offload of five UAVs. By dynamically partitioning the network, the hierarchical framework preserves the integrity of the HAPS links for mission-critical nodes while seamlessly shifting delay-tolerant traffic to the ground infrastructure, avoiding significant packet drops without requiring computationally heavy gradient recalculations.

\subsection{Training-Phase Convergence and Multi-Objective Analysis}

\begin{figure*}[t]
    \centering
    % Row 1
    % \begin{subfigure}{0.48\linewidth}
    %     \centering
    %     \includegraphics[width=\linewidth]{images/binned_metrics_outputs/decoded_total_reward.pdf}
    %     \caption{Total Reward ($\sum \vec{\mathcal{R}}_t^m$)}
    %     \label{fig:total_reward}
    % \end{subfigure}
   
    \begin{subfigure}{0.48\linewidth}
        \centering
        \includegraphics[width=\linewidth]{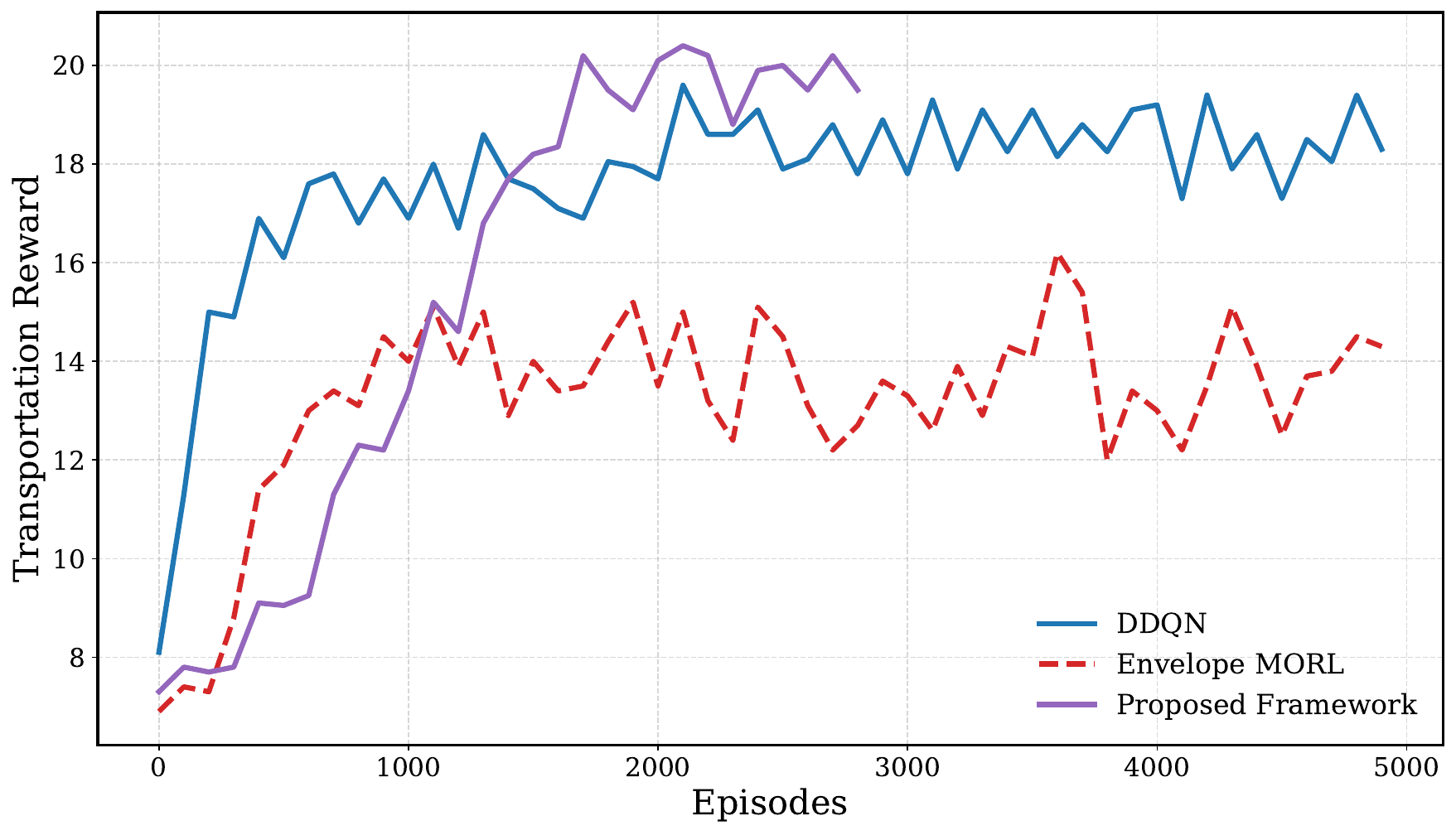}
        \caption{Transportation Efficiency ($R_{\text{tran},t}^m$)}
        \label{fig:tran_reward}
    \end{subfigure}
 \hfill
    \begin{subfigure}{0.48\linewidth}
        \centering
        \includegraphics[width=\linewidth]{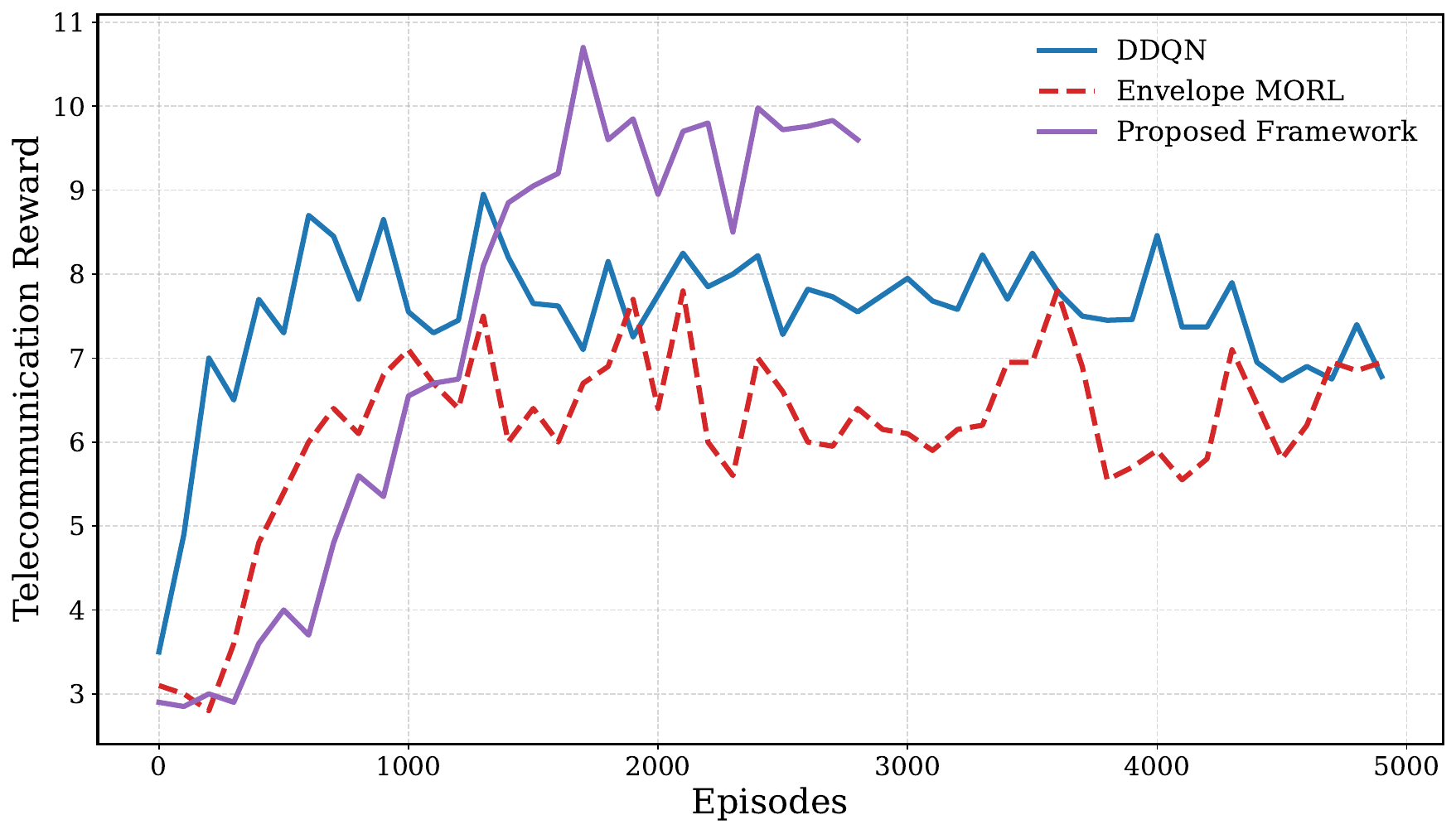}
        \caption{Telecommunication Utility ($R_{\text{tele},t}^m$)}
        \label{fig:tele_reward}
    \end{subfigure}

    % \vspace{0.5em}

    % Row 2
    % \begin{subfigure}{0.48\linewidth}
    %     \centering
    %     \includegraphics[width=\linewidth]{images/binned_metrics_outputs/decoded_telecommunication_reward.pdf}
    %     \caption{Telecommunication Utility ($R_{\text{tele},t}^m$)}
    %     \label{fig:tele_reward}
    % \end{subfigure}
    % \hfill
    % \begin{subfigure}{0.48\linewidth}
    %     \centering
    %     \includegraphics[width=\linewidth]{images/binned_metrics_outputs/decoded_step_count.pdf}
    %     \caption{Episode Length (Survival)}
    %     \label{fig:step_count}
    % \end{subfigure}

    \caption{Training phase convergence comparison of the proposed hierarchical generative AI framework.
    %\textcolor{red}{Please harmonize how you call your proposed solution. Here it is LLM-LLM Dual Agent, and below it is Proposed Framework. Please make up your mind over ALL Figures and for all Methods!}. {\color{purple} Zijiang comment: I update the legend to be consistent with Proposed Framework}
    }
    \label{fig:multi_metric_comparison}
\end{figure*}

Fig.~\ref{fig:multi_metric_comparison} illustrates the learning curves of the evaluated algorithms. 
%The empirical results confirm that the proposed hierarchical framework significantly outperforms all benchmark methods across the multi-objective reward vectors.
%\textbf{(a)~Total Reward:} 
% As shown in Fig.~\ref{fig:total_reward}, while DDQN exhibits rapid initial learning, it quickly becomes trapped in a suboptimal local minimum, fluctuating around a reward of 25. In contrast, the proposed architecture exhibits a brief cognitive exploration phase before accelerating, overtaking the DDQN near episode 1,500, and stabilizing at a peak reward of nearly 30. In contrast, Envelope-MORL \cite{yan2024generalized} struggles with the highly coupled dynamics, saturating at a reward close to 20, due to its inability to dynamically adapt its scalarization weights.
%\textbf{(b)~Transportation Efficiency ($R_{\text{tran},t}^m$):} 
Fig.~\ref{fig:tran_reward} reflects the results of {$\sum \vec{\mathcal{R}}_t^m$} in the physical domain. after convergence beyond 2000 episodes, validating the LLM's strong semantic reasoning for assertive, goal-oriented spatial maneuvers. Conversely, the conventional DDQN agent converges towards a conservative, sub-optimal policy that artificially minimizes safety penalties.

%\textbf{(c)~Telecommunication Utility ($R_{\text{tele},t}^m$):} 
Fig.~\ref{fig:tele_reward} presents the telecommunication utility as a function of the number of RL episodes. % highlights the superiority of LLM-driven network orchestration. 
Although DDQN initially saturates at a reward of approximately 7.5, the hierarchical LLM framework outperforms it after episode 1,500 towards a reward of nearly 10. By decoupling the fast-timescale DDQN channel selection from the slow-timescale LLM flight planning, the dual-agent proactively anticipates signal degradation and initiates timely handovers.

\subsection{Scalability under Varying 3D Traffic Density}
\begin{figure*}[t]
    \centering
    \includegraphics[width=\linewidth]{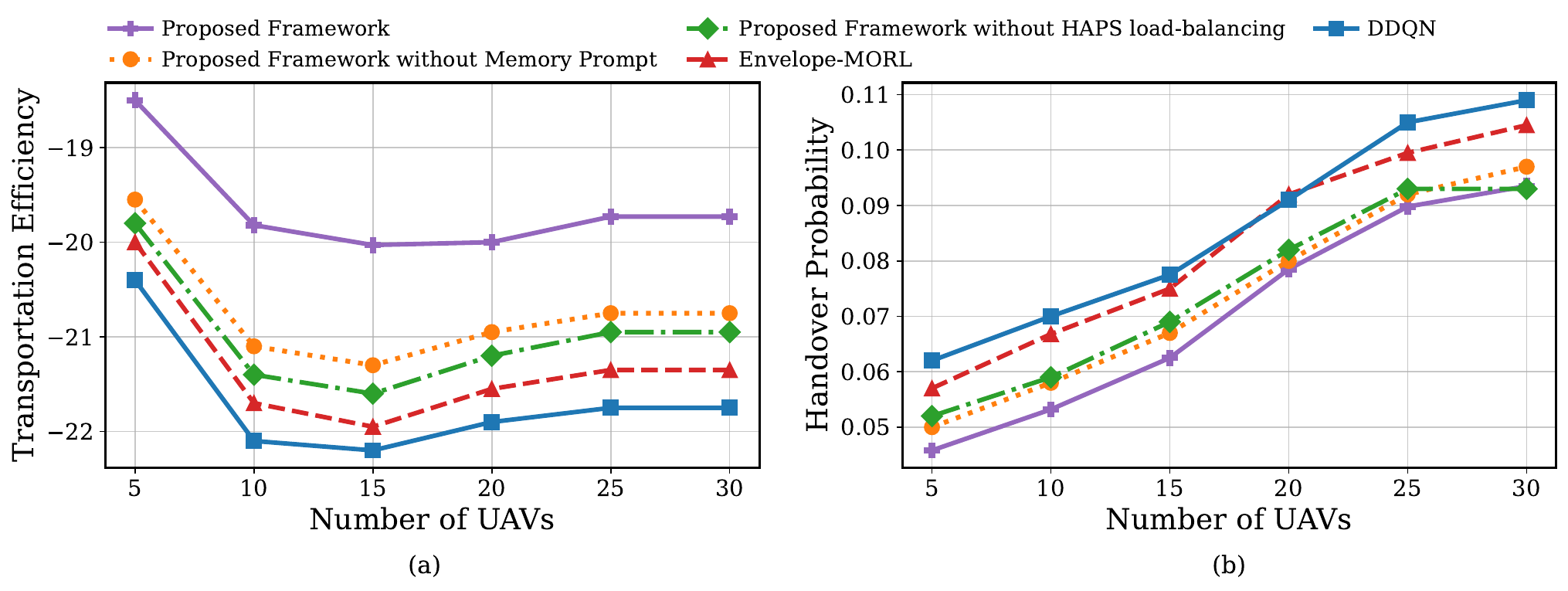}
    \caption{Scalability evaluation comparing operational penalties and safety metrics under varying 3D UAV traffic densities.}
    \label{fig:evaluation_comparison}
\end{figure*}
{{To evaluate system robustness and the contribution of individual framework components, Fig.~\ref{fig:evaluation_comparison} presents key operational metrics averaged over 500 evaluation episodes as the number of active UAVs ($M$) scales from 5 to 30. In addition of the proposed framework, we evaluate the performances of two degraded variants of the proposed framwork, namely:
\begin{enumerate}
    \item {w/o Memory Prompt ($\mathcal{P}^{\mathrm{mem}}_t$):} The UAV edge-agents rely solely on their base state observations ($\mathbf{p}^{\text{dyn}}_t$) without retrieving historical reflection summaries ($c_t^m$).
    \item {w/o Meta-Controller:} The global \texttt{Qwen3.5-122B} meta-controller is disabled. UAVs execute greedy network associations without HAPS capacity gating ($C_{\max}^{\mathrm{HAPS}}$).
\end{enumerate}
As $M$ increases, maintaining efficient flight paths becomes increasingly difficult due to higher interference and required collision avoidance maneuvers. As illustrated by the transportation efficiency trends in Fig.~\ref{fig:evaluation_comparison}a, the transportation reward of the DDQN baseline degrades severely as $M$ scales towards 30. In contrast, the proposed framework achieves the best transportation reward (even without memory prompt or HAPS load-balancing), showcasing a superior and stable trajectory profile even at maximal $M=30$. The envelope-MORL benchmark achieves better performance than DDQN but worse than the proposed framework and its variants. Finally, for all approaches, as $M$ grows from 5 to 15, the transportation reward degrades {due to {the rapid onset of spatial congestion, forcing UAVs to deviate constantly from their optimal forward trajectories and execute erratic evasive movements.}However, when $M$ exceeds 15, the reward slightly increases then saturates for higher $M$ values. This is due to {the emergence of collective swarm alignment: at high densities, the airspace becomes fully saturated, forcing the agents to adopt uniform, platooning to minimize continuous collision penalties. }}
Finally, Fig.~\ref{fig:evaluation_comparison}b evaluates network stability by plotting the probability of vertical handovers. DDQN exhibits the highest handover probability for any $M$ value, reacting to localized channel fading with uncoordinated network switches. Envelope-MORL performs marginally better. In contrast, the proposed framework maintains the lowest and most stable handover probability. The performance slightly degrades when the HAPS meta-controller or Memory Prompt is removed, highliting their important role in executing vertical handovers only when necessary. 
%The poor performance of the without Meta-Controller baseline further confirms that the explicit capacity gating of the HAPS meta-controller is required to successfully filter out transient signal fluctuations and execute vertical handovers only when strategically necessary.
}
}

\subsection{Impact of LLM Scale and Memory Retrieval}
\label{subsec:ablation_study}
%To rigorously justify the architectural complexity of the proposed framework, we conduct an ablation study isolating the contributions of the semantic memory pool and the dual-tier LLM hierarchy. 
\begin{table}[t]
    \caption{Ablation Study: Impact of Hierarchical Components on System Performance ($M=30$ UAVs)}
    \centering
    \scriptsize
    \renewcommand{\arraystretch}{1.3} % Adds padding for readability
    \begin{tabularx}{\columnwidth}{@{}l c c X X X@{}}
        \toprule
        \multirow{2}{*}{\textbf{Architecture Variant}} & 
        \multicolumn{2}{c}{\textbf{Active Components}} & 
        \multicolumn{3}{c}{\textbf{Performance Metrics}} \\
        \cmidrule(lr){2-3} \cmidrule(l){4-6}
        & \textbf{HAPS LLM} & \textbf{Memory} & \textbf{Collision Rate ($C_{\text{safe}}$)} & \textbf{Avg. Datarate (Gbps)} & \textbf{Capacity Violations} \\
        \midrule
        \textbf{Proposed Framework} & \checkmark (\texttt{122B}) & \checkmark & \textbf{0.07} & \textbf{1.18} & \textbf{0.0\%} \\
        w/o Memory Prompt & \checkmark (\texttt{122B}) & $\times$ & 0.19 & 1.15 & 0.0\% \\
        w/o Meta-Controller & $\times$ (Greedy) & \checkmark & 0.08 & 0.62 & 84.5\% \\
        \bottomrule
    \end{tabularx}
    \label{tab:ablation_study}
\end{table}
%The results are presented in {Fig.~\ref{fig:evaluation_comparison} and}
In Table~\ref{tab:ablation_study}, we present the comparative analysis of the proposed framework against its ablation variants in terms of complexity, collision rate, average datarate, and capacity violations. Removing the episodic memory retrieval ({w/o Memory Prompt}) causes a severe degradation in physical safety (high collision rate). Indeed, without access to past cognitive reflections, the \texttt{Qwen3.5-9B} edge-agents lose their zero-shot spatial awareness, causing the collision rate to spike from $0.07$ to $0.19$. This proves that while the LLM possesses inherent reasoning capabilities, dynamic grounding via $\mathcal{P}^{\mathrm{mem}}_t$ is necessary to bridge the gap to continuous control.
On the other hand, disabling the strategic HAPS orchestrator ({w/o Meta-Controller}) isolates the failure to the telecommunications domain. Without the \texttt{Qwen3.5-122B} model anticipating congestion, the UAVs greedily associate with the HAPS, violating the $C_{\max}^{\mathrm{HAPS}}$ threshold in $84.5\%$ of the time steps. This resource starvation causes severe packet drops, degrading the aggregate network throughput to $0.62$~Gbps.

\section{Conclusion and Future Work}
\label{sec:conclusion}

As UAVs increasingly populate the 3D aerial highway, balancing aggressive physical mobility with reliable mission-critical connectivity remains a challenge. This paper resolves this transportation-telecommunication conflict by proposing a novel, generative AI-driven hierarchical cognitive framework for ITNTN-enabled UAV swarms. By formulating the environment as an H-MO-POMDP, we successfully distributed decision-making across two highly coordinated tiers. At the strategic level, a slow-timescale HAPS Meta-Controller utilizes global cognitive reflection to execute dynamic network load balancing, effectively mitigating capacity saturation. At the tactical edge level, we pioneered a hybrid dual-agent architecture that explicitly bridges the abstraction gap between semantic AI and continuous physics.
%; an onboard LLM handles spatial reasoning and collision avoidance, while a deterministic motion decoder and a parallel DDQN handle high-frequency rotor kinematics and U2I handovers. 
Extensive 3D simulations demonstrate that the proposed framework outperforms baseline DRL and MORL methods in terms of transportation efficiency, telecommunication data rate, collision rate, and HO events. 
%By leveraging few-shot memory retrieval and task discretization, the hierarchical LLM approach achieves zero-shot safety awareness, significantly suppressing 3D collision rates in ultra-dense traffic scenarios while maintaining high-throughput, low-latency telecommunication links.


% % -------------------------------------------------------------------------
% % APPENDIX
% % -------------------------------------------------------------------------
% % \appendices

% % \section{Supplementary Prompts and LLM Interaction Examples}
% % (Your appendix content unchanged, but formatted for IEEE.)

% % -------------------------------------------------------------------------
% % REFERENCES
% % -------------------------------------------------------------------------
\bibliographystyle{IEEEtran}
\bibliography{main-clean}

\end{document}